# Human and AI collaboration for pulmonary nodule segmentation

Hongqiao Dong[1,7,*], Wenhao Chi[5,1,*], Ruobing Liang[1,7,*], Xiaokui Yang[4], Wenhua Liang[3,6], Peng Hou[2], Wenjun Pu[2], Yipeng Zhao[1], Ping Chen[2], Haiping Liu[2,8,#], Jianxing He[3,6,#], Bo Liu[1,#;]

*Contributed equally.

#Corresponding authors: Jianxing He, drjianxing.he@gmail.com; Haiping Liu, lhp586@163.com; Bo Liu, bliu@amss.ac.cn.

[1]State Key Laboratory of Mathematical Sciences, Academy of Mathematics and Systems Science, Chinese Academy of Sciences, Beijing, China

[2]PET/CT Center, The First Affiliated Hospital of Guangzhou Medical University, Guangzhou, China

[3]Department of Thoracic Surgery and Oncology, The First Affiliated Hospital of Guangzhou Medical University, Guangzhou, China

[4]Department of Mathematical Science, Tsinghua University, Beijing, China

[5]Yau Mathematical Sciences Center, Tsinghua University, Beijing, China

[6]China State Key Laboratory of Respiratory Disease & National Clinical Research Centre for Respiratory Disease, Guangzhou, China

[7]School of Mathematical Sciences, University of Chinese Academy of Sciences, Beijing, China

[8]National Center for Respiratory Medicine, National Clinical Research Center for Respiratory Disease, Guangzhou Institute of Respiratory Health, The First Affiliated Hospital of Guangzhou Medical University, Guangzhou, China

## Abstract

Medical expert annotators are scarce, and blind reliance on artificial intelligence (AI) can be misleading, motivating approaches in which humans, particularly junior medical trainees or even non-medical personnel, collaborate with AI to achieve robust medical segmentation. Although the Segment Anything Model (SAM) shows promise for general-purpose image segmentation, its performance in human-AI collaboration for specialized medical tasks has not been thoroughly evaluated. Here we present Hi-Seg, a human-in-the-loop segmentation framework for pulmonary nodules built on SAM. Humans iteratively refine prompts through trial-and-error learning and semantic reasoning, progressively guiding SAM toward higher-quality masks. Using chest CT scans from 1,179 patients across 12 centers, we conducted the first large-scale external validation of collaborative human-SAM segmentation. Across all annotator groups, Hi-Seg achieved a mean Dice score of almost 85%, outperforming five state-of-the-art deep learning models by 10-22% and 13 SAM variants by 1-29%. Hi-Seg improved segmentation accuracy while reducing annotation time for medical annotators, and briefly trained non-medical annotators achieved performance comparable to that of the junior medical student. These findings suggest that human-in-the-loop segmentation can reduce clinician workload, enable scalable crowdsourced annotation, and transform clinical workflows by facilitating the safe and efficient integration of foundation models into routine clinical practice.

## Introduction

This study stems from a critical yet often overlooked observation: despite extensive research on deep learning for pulmonary nodule management, high-quality segmentation masks that are essential for training data-intensive and generalizable artificial intelligence (AI) models remain scarce. Our initial approach involved recruiting radiologists to manually annotate nodules. However, given their limited availability and heavy clinical workload, this approach proved to be infeasible and inefficient. This challenge raised a central question: how could we rapidly generate high-quality lesion segmentation masks to support robust AI training?

AI-driven tools are increasingly used to improve the efficiency of medical image segmentation. Deep learning models[1-3] and, more recently, prompt-driven foundation models[4-7] have greatly accelerated segmentation, yet achieving stable and accurate results remains challenging with these approaches alone. As a result, extensive clinician involvement for review and correction is still required.

Deep learning-based segmentation performance largely depends on the model's ability to learn discriminative features that robustly distinguish abnormalities from surrounding tissues[8]. Model performance varies substantially with lesion appearance. Solid nodules, which exhibit high-density contrast against lung parenchyma, are segmented accurately (mean Dice score: 81%), whereas ground-glass opacities, characterized by subtle haze-like attenuation with texture resembling normal parenchyma, remain challenging, yielding lower accuracy (mean Dice score: 67%)[9]. These disparities reveal the inability of fully automated deep learning models to reliably segment lesions with texture similar to healthy tissue[10].

The segmentation accuracy of foundation models is highly sensitive to the quality of prompts. These models generate an initial mask from prompts and refine it iteratively using additional prompts. Poorly designed prompts could lead to performance deterioration or even complete failure. For instance, the Segment Anything Model (SAM)[4] achieved robust segmentation of brain tumors on MRI scans and polyps on colonoscopy images when provided with prompts derived from ground-truth masks[6]. However, without such ideal prompts, performance dropped sharply, for example, to a Dice score of only 42% for lung nodule segmentation across 50 chest CT images[7]. Nearly all reported high-performance results for SAM have relied on prompts automatically generated from ground-truth masks, a condition rarely met in clinical practice.

The dilemma is evident: blind reliance on AI can be misleading, and expert annotators are scarce everywhere. Human and AI collaboration offers transformative opportunities in clinical practice[11,12]; nevertheless, the efficacy of foundation models such as SAM in collaborating with humans on specialized medical segmentation tasks remains underexplored. This motivates approaches that enable humans, particularly junior trainees or even briefly trained non-medical personnel, to work with SAM to achieve robust segmentation.

Here, we present Hi-Seg, a human-in-the-loop segmentation tool for medical imaging built on SAM (Fig. 1a). In Hi-Seg, SAM generates initial candidate masks based on human-provided prompts, which annotators iteratively refine by adding new prompts. By observing SAM's feedback, annotators learn which prompts yield high-quality masks and which do not, gradually understanding how prompt placement and type influence segmentation outcomes. This iterative, closed-loop interaction enables informed prompt selection and accelerates convergence to accurate masks. Unlike fully automated AI approaches, Hi-Seg balances efficiency and precision by integrating human oversight and semantic reasoning to guide SAM.

## Results

### Dataset characteristics

We comprehensively evaluated Hi-Seg for pulmonary nodule segmentation, a task complicated by substantial variability in nodule size, morphology, density, and anatomical location (Fig. 1h, i, Supplementary Table 1). The publicly available LIDC-IDRI chest CT dataset[13] (1,010 patients) was used for training and internal validation. External testing was performed on two independent datasets: a single-center cohort of 904 patients and a multi-center cohort of 275 patients from 12 institutions[14,15] (Fig. 1e, Extended Data Fig. 1, Methods).

All ground-truth segmentation masks were annotated by radiologists following a standardized workflow (Extended Data Fig. 2). In the LIDC-IDRI dataset, each nodule was independently delineated by at least three thoracic radiologists, and voxels marked by at least two radiologists were retained. The largest connected component was extracted to form a 3D mask, which was subsequently resampled to a standard resolution. For the single-center and multi-center datasets, a radiologist with 12 years of experience identified the nodule center, after which preliminary 3D masks were generated using a combination of U-Net[16], active contour methods[17], and morphological operations. The annotation effort, conducted by the radiologist over nearly three years (2020-2022), was labor-intensive, motivating us to look for ways to accelerate segmentation and help more patients. From each 3D mask, 256×256 2D slices were extracted along the axial, coronal, and sagittal planes passing through the mask centroid to serve as ground-truth annotations.

The distribution of nodule characteristics differed across datasets, with potential affection on model generalizability. Details on how these characteristics were extracted are provided in the Methods. We observed substantial variation in nodule size, density, contour, and internal features across datasets (Extended Data Fig. 2, Supplementary Table 1). For example, the proportion of spiculated nodules was 24.5% and 20.0% in the training and internal validation sets, respectively, but increased to 71.6% and 60.0% in the single-center and multi-center external test datasets. Because spiculation increases segmentation complexity, such differences might contribute to performance discrepancies across datasets.

**Human-in-the-loop segmentation for pulmonary nodules with SAM (Hi-Seg)**

In Hi-Seg, human annotators iteratively interact with SAM to refine segmentation masks (Fig. 1a). Preprocessed chest CT images are presented sequentially in the axial, coronal, and sagittal views (Extended Data Fig. 3), mirroring routine clinical workflows. Annotators identify and segment nodules across these views in a stepwise manner. To initiate segmentation, the annotator left-clicks within the nodule region to indicate the target location, prompting SAM to generate a pixel-level mask. If the mask is unsatisfactory, the annotator refines it further: (i) if the mask fails to fully encompass the nodule, a left-click on an uncovered region instructs SAM to include it; and (ii) if the mask spills into the background, a right-click on the extraneous background instructs SAM to exclude it. SAM updates the mask based on the new prompts and the previous prediction, and this process repeats until the annotator is satisfied. SAM generates image embeddings using a Vision Transformer (ViT) encoder[18], a computationally intensive step. By contrast, Hi-Seg pre-computes and reuses these embeddings, enabling low-latency, human-in-the-loop segmentation (Methods).

We recruited five participants with diverse backgrounds (Fig. 1f), including a senior radiologist with 12 years of experience in respiratory diseases, a 5th-year junior medical student, and three non-medical annotators. The senior radiologist delivered a 1.5-hour training session on lung nodule CT interpretation to the non-medical annotators. Following training, the non-medical annotators independently localized and segmented nodules in the Hi-Seg interface (Fig. 1g), with boundary quality assessed by the radiologist. The interface then presented 30 two-dimensional CT images from 10 patients in the LIDC-IDRI dataset, allowing all participants to practice human-in-the-loop segmentation, with an emphasis on achieving high-quality masks using minimal prompts (see Methods).

**Performance evaluation and comparison**

*Segmentation performance of Hi-Seg versus deep learning models.* Hi-Seg was first compared with five deep learning architectures (U-Net[19], U-Net++[20], Attention U-Net[21], TransU-Net[22], and nnU-Net[8]) on independent single-center and multi-center external test datasets. All deep learning models were trained on the LIDC-IDRI dataset, and the best-performing variant from each architecture, selected based on internal-validation Dice score, was used for external testing (Methods, Fig. 1d). Internal validation showed performance consistent with prior reports[23] (Supplementary Table 2). In a collaborative setting involving five participants, Hi-Seg achieved mean Dice scores of 84.47% (95% CI: 84.05-84.90%) and 87.00% (95% CI: 86.02-87.98%) on single-center and multi-center datasets, respectively, outperforming all deep learning models (all $P < 0.001$) (Table 1, Supplementary Table 3, Fig. 2). These findings demonstrate the efficacy of human and SAM collaboration for pulmonary nodule segmentation.

*Comparison of Hi-Seg with SAM-based models using pseudo-human prompts.* We compared Hi-Seg, which uses real human interactive prompts, with 13 previously reported SAM-based models that instead relied on pseudo-human prompts derived from ground-truth masks. SAM[4], HQ-SAM[24], and MedSAM2[25] were each paired with four pseudo-human prompts (#1-#4, Methods, Fig. 1c) to generate 12 models, and MedSAM[7] was additionally tested with bounding box prompts. HQ-SAM was fine-tuned on 44,320 natural images; MedSAM on 1,570,263 medical images across 10 modalities and 30 cancer types; and MedSAM2 is a refined SAM2[26] variant on 78 multimodal medical datasets. Hi-Seg prompted by human annotators consistently outperformed all SAM-based models using pseudo-human prompts (all $P < 0.001$) (Table 1, Supplementary Table 3, Fig. 2a, b). The best-performing pseudo-human model, MedSAM2 with prompt #4, achieved Dice scores of 83.08% (95% CI: 82.56-83.60%) and 83.99% (95% CI: 82.96-85.02%), yet remained inferior to Hi-Seg. These results reveal the advantages of real human prompting, which carries implicit semantic reasoning, and the limitations of pseudo-human prompts in approximating human decision-making for prompt placement.

*Subgroup performance across nodule characteristics.* The above analyses were based on an overall evaluation across all nodule characteristics. We next assessed segmentation performance within individual nodule subgroups, stratified by nodule characteristics, for annotators with varying levels of medical expertise under the human-in-the-loop paradigm (Hi-Seg). Performance was compared with manual delineation alone (the common clinical scenario) and with fully automated approaches, including deep learning models, SAM-based models using pseudo-human prompts, and prompt-free SAM[4]. Across nearly all nodule subgroups, Hi-Seg achieved the highest segmentation accuracy (Fig. 2c, Supplementary Table 4), with performance gains evident even in the most challenging cases, including small nodules (<15 mm) and nodules with pleural tags. A consistent performance hierarchy was observed across subgroups, with Hi-Seg performing best, followed by manual delineation, pseudo-human-prompted SAMs, deep learning models, and prompt-free SAM. Overall, Hi-Seg demonstrated robust performance across nodules with diverse characteristics.

*Contribution of human prompting and iterative feedback refinement to segmentation performance.* We performed ablation experiments to disentangle the respective contributions of human prompting, iterative interaction, and feedback-based refinement to improvements in segmentation performance (Table 2). Prompt-free SAM, which generates masks in a single pass without any prompts, exhibited poor performance, achieving a Dice score of only 45.27% across two external test datasets. Introducing real human or pseudo-human prompts yielded substantial gains even without iteration: one-shot Hi-Seg and one-shot SAM increased the Dice score to 65.27% and 65.41%, respectively. Building on this, pseudo-human-prompted SAMs further benefited from iterative prompting, reaching a Dice score of 75.44%; however, performance remained lower than that of the full Hi-Seg framework, which combines iterative, feedback-driven human prompt adjustment and achieved the highest Dice score of 84.81%. Overall, these ablation results indicate that the primary performance gains arise not merely from human prompting itself, but from a human-in-the-loop workflow. In this workflow, iterative feedback enables continuous error correction, knowledge accumulation, and progressive mask refinement, effectively integrating human conceptual understanding and semantic reasoning with the model's strength in rapid boundary delineation.

*Hi-Seg narrows expertise-dependent performance gaps while improving annotation efficiency*. Hi-Seg improved segmentation accuracy across all annotator groups, with gains varying by expertise level and substantially narrowing the performance gap relative to the senior radiologist (Table 3, Fig. 2c, d, Supplementary Table 5). The largest improvements were observed for non-medical annotators (3.80% increase over manual annotation), whereas improvements, although smaller, remained statistically significant for the senior radiologist (1.43% over manual; $P < 0.001$) and the junior medical student (1.19% over manual; $P < 0.001$). Under Hi-Seg, the medical student achieved accuracy comparable to the senior radiologist's manual performance, while non-medical annotators reached accuracy levels similar to the medical student's manual annotations. In terms of efficiency, Hi-Seg reduced annotation time by approximately 30% for both the medical student and non-medical annotators. By contrast, a modest decrease in efficiency was observed for the senior radiologist, likely reflecting ceiling effects in expert manual performance and the additional cognitive demands associated with transitioning from an observe-and-delineate workflow to a prompt-generate-refine interaction paradigm. Overall, Hi-Seg reduces dependence on expert-level annotation, supporting its potential utility for crowdsourced medical image segmentation across annotators with diverse levels of expertise.

*Segmentation failure patterns across nodule subgroups define the boundaries of human-AI collaboration*. Our analyses identified nodule subgroups in which non-medical annotators and fully automated models were particularly prone to segmentation failure, pinpointing cases that still require expert oversight to ensure patient safety (Supplementary Table 6). Failure rates were highest for ground-glass nodules, owing to their subtle, ill-defined radiological appearance and attenuation values closely resembling adjacent normal lung parenchyma, which challenge inexperienced annotators and automated models with limited discriminative capacity. By contrast, the senior radiologist and medical student accurately delineated these lesions using contextual and anatomical cues, rendering AI assistance largely unnecessary in this subgroup (Supplementary Table 4). Collectively, these findings demonstrate that the radiological complexity of certain nodule subgroups constrains the scope of human-AI collaboration and emphasize the need for expert involvement to maintain segmentation reliability and clinical safety in high-risk cases.

*Interactive feedback guides informed prompt selection and accelerates convergence to expert-level segmentation*. A case study illustrates how interactive feedback enables annotators to iteratively refine segmentation masks and rapidly converge toward the ground truth. A senior radiologist, assisted by Hi-Seg, performed seven independent segmentation attempts on the same CT scan (Fig. 3). The initial prediction overestimated the target nodule and required multiple prompts to remove redundant regions. As prompting strategies were progressively refined and prompt points placed closer to the nodule boundary, mask quality improved rapidly: by the sixth and seventh attempts, accurate segmentation was achieved with only three prompts, compared with 14 prompts in the first attempt. Extended Data Figs. 4-7 further show that repeated interaction allowed annotators to internalize how prompt placement affects predicted masks across different nodule subtypes. This feedback-driven learning enabled more informed prompt selection, reducing interaction burden while yielding masks comparable to the ground truth, consistent with the ablation results.

Across the extensive and multifaceted comparative experiments described above, the human-in-the-loop segmentation paradigm (Hi-Seg) achieved the highest segmentation accuracy and efficiency. This performance is attributable to its interactive feedback mechanism, which supports trial-and-error learning during human-AI collaboration and progressively refines annotators' prompting strategies. Through repeated interaction, annotators reinforce prompting behaviors that yield high-quality segmentation outcomes while discarding those that produce low-quality results, thereby enabling efficient human-AI collaboration and improved overall annotation performance.

## Discussion

In cancer care, medical image segmentation is a critical prerequisite for a wide range of time-sensitive, life-saving medical procedures[27], where the segmentation speed and accuracy directly impact treatment outcomes and, consequently, patient safety[28-33]. For example, in prostate cancer radiotherapy, even small contouring errors, on the order of 1 mm relative to the reference, have been shown to increase recurrence risk by 8-24%[32].

Manual annotation by experienced clinicians remains the clinical gold standard[34,35]. However, its inherently time-consuming and labor-intensive nature not only causes clinician burnout but may also delay subsequent patient care[36]. The surge in demand for medical image segmentation, coupled with the scarcity of medical experts with high annotation experience[37], further exacerbates this challenge.

Although artificial intelligence-based segmentation tools have demonstrated strong performance across multiple imaging modalities and lesion types, existing automated approaches, whether task-specific deep learning models or general-purpose image segmentation foundation models, continue to suffer from declines in generalization when deployed in real-world clinical settings. This limitation largely arises from distributional heterogeneity between training data and clinical data, as well as the reliance of foundation models on idealized, non-clinically realistic prompts. Together, these factors constrain the reliable deployment of automated segmentation methods in clinical practice.

When applied to medical image segmentation, the Segment Anything Model (SAM) typically relies on prompts derived from ground-truth nodule masks (hereafter referred to as pseudo-human prompts) [5-7,38]. However, in external datasets and real-world clinical settings, such lesion masks are typically unavailable a priori, resulting in performance degradation. This limitation is particularly pronounced in pulmonary nodule segmentation, where large heterogeneity in nodule size, shape, density, and anatomical location poses a major challenge for SAM when idealized prompts are absent. For example, on the LUNA16 dataset (888 CT scans comprising 1,186 nodules), Dice scores under five pseudo-human prompt strategies ranged from 64.86% to 84.87%[6]. MedSAM[1] using pseudo-human prompts exhibited performance drops on external datasets, with Dice scores of 74.8% on 50 chest CT scans[7] and 28.3% on 769 CT scans from 200 patients with non-small cell lung cancer[39]. Similarly, another SAM variant, MedSegX, achieved a Dice score of 60.98%[37] on the same 769-scan dataset[39], indicating that overall segmentation accuracy remains insufficient to meet the clinical

[1] MedSAM achieved high performance on an internal validation set (146 chest CTs; Dice 93.8%).

demands.

Our proposed human-in-the-loop segmentation method (Hi-Seg) relies exclusively on real human prompts. During a feedback-driven interactive process, annotators iteratively learn and refine their prompting strategies through trial and error, quickly converging on high-quality masks. Effective prompting behaviors that yield high-quality masks are reinforced, whereas inefficient strategies are discarded. This trial-and-error learning process requires only 90 minutes of practicing segmentation 30 CT images under Hi-Seg. Leveraging this feedback-driven refinement, Hi-Seg demonstrates robust performance across both large-scale single-center and multi-center external test sets, achieving Dice scores of 86.5% and 88.3%, respectively. In contrast, existing SAM-based methods that generate pseudo-human prompts from ground-truth masks lack a feedback-learning mechanism and cannot progressively adjust prompting strategies. These results indicate that real human-generated prompts better align with clinical workflows and exhibit better generalizability.

In the existing literature, only three studies have evaluated clinician-prompted SAM, primarily focusing on improvements in annotation efficiency[2]. Across segmentation tasks, including adrenocortical carcinoma[7], optic cup[38], and general medical targets[6], clinician-prompted SAM substantially reduced manual annotation time. However, these studies did not investigate whether clinicians can progressively learn and develop effective prompting strategies through interactive feedback, nor did they address pulmonary nodule segmentation. Moreover, the clinicians involved were relatively homogeneous in professional background and experience, with a median of 10 years in practice (range, 6-10 years). In contrast, our proposed human-in-the-loop segmentation framework (Hi-Seg) introduces an iterative, feedback-driven refinement mechanism that enables annotators to gradually understand the mapping between prompts and model predictions through trial-and-error interactions, continuously improving prompting strategies for more efficient human-AI collaboration. This study also includes annotators with more diverse medical backgrounds, including non-medical individuals, to comprehensively assess the applicability and generalizability of the human-in-the-loop segmentation paradigm.

---

[2] Two radiologists with six and eight years of abdominal imaging experience used MedSAM to segment 10 adrenocortical carcinoma cases (733 CT slices), reporting an 83% reduction in annotation time compared with manual segmentation[6]. Three ophthalmologists with over ten years of experience applied Med-SA, a SAM variant with lightweight encoder adapter, to segment optic cups in fundus images[36]. Three clinicians with 10 years of experience annotated 100 images (covering nine modalities, 55 target classes, and 620 masks) from the COSMOS 1050K dataset using SAM, achieving an approximately 30% reduction in annotation time relative to manual annotation[5].

During human-in-the-loop interaction, annotators progressively develop efficient prompting strategies grounded in semantic reasoning that are not inherently available to existing automated AI methods. Leveraging domain knowledge of nodule radiologic characteristics, annotators can identify regions requiring correction and determine more appropriate prompt placements accordingly. For example, in Extended Data Fig. 4, the annotator recognized a pleura-attached nodule and, in response to SAM's tendency to generate overly extended masks around the adhesion region that erroneously encompassed adjacent lung parenchyma, positioned the prompt near the adhesion site to effectively constrain the segmentation boundary. The incorporation of such semantic concepts into interactive prompting not only enhances the interpretability of human prompting behavior but also improves the effectiveness of human-AI collaboration.

Because cancer imaging analysis heavily relies on specialized medical knowledge, prior studies have generally avoided crowdsourcing lesion annotations[40-42]. For instance, in lung cancer, manual annotation of tumors and lymph nodes for a single patient typically requires 15-90 minutes by an experienced clinicians[43]. Our study demonstrates that the proposed Hi-Seg framework enables non-medical annotators to effectively participate in large-scale nodule segmentation. Through iterative human-in-the-loop learning, these non-medical annotators progressively develop prompting strategies comparable to those of radiologists, producing expert-level segmentation masks. These findings support a hierarchical annotation workflow in which non-medical annotators generate preliminary segmentations via Hi-Seg, followed by radiologist supervision and refinement of complex cases, such as ground-glass nodules, thereby improving clinical annotation efficiency.

This study has several limitations. First, segmentation was limited to 2D CT images in three views; future work should explore 3D interactive segmentation between humans and foundation model. Second, the number of human annotators was relatively small, although it remains the largest cohort among comparable studies[6,7,38]. Recruiting a larger and more diverse cohort with varying levels of expertise could further reduce bias. Third, we focused on short-term interactive learning; investigation of long-term effects across diverse imaging tasks would be worthwhile. Finally, although this study focused on generating high-quality segmentation masks for data-rich AI models, future applications in diverse clinical settings may be possible, for example, enabling accurate calculation of nodule volume-doubling time to assess malignancy risk[44] or precise delineation of radiotherapy target volumes to reduce recurrence risk while protecting organs at risk[30,33].

Medical expert annotators are scarce, and blind reliance on artificial intelligence can be misleading. We present a human-centered AI framework[45], SAM-based human-in-the-loop segmentation (Hi-Seg), integrating annotators from diverse medical backgrounds, including both medical and non-medical personnel. Large-scale, multicenter testing demonstrates that Hi-Seg outperforms fully automated AI approaches, including state-of-the-art deep learning models and fine-tuned pseudo-human-prompt SAM variants. This approach improves segmentation accuracy across all annotator groups while reducing annotation time. Through iterative trial-and-error and semantic reasoning, annotators progressively guide SAM to generate high-quality masks. These findings support integrating foundation models into clinical workflows, establishing a generalizable framework for collaborative segmentation in oncology: even junior medical students and non-medical annotators could work with the model, providing clinicians with a second, safe pair of hands. Although widespread deployment of foundation models may raise economic and power-concentration concerns[46], human-in-the-loop schemes could facilitate broader dissemination of medical AI, enhance healthcare accessibility, and reduce the cancer burden[47], while mitigating disparities in cancer treatment outcomes both globally and within countries[48].

**Acknowledgements**

This study utilized a multi-center dataset established under the leadership of the late Professor Jiahe Tian, former Director of the Department of Nuclear Medicine, Chinese PLA General Hospital, Beijing. We remain deeply grateful for his foundational contributions. This study has benefited immensely from extensive discussions with Emeritus Professor Michiel Keyzer (B.L.'s postdoctoral supervisor, SOW-VU, Vrije Universiteit Amsterdam, The Netherlands) and from three rounds of his rigorous and invaluable critique of the full manuscript. All of his comments and criticisms have been carefully addressed and incorporated, for which the authors are deeply grateful. We also thank Associate Professor Lia van Wesenbeeck (SOW-VU, Vrije Universiteit Amsterdam, The Netherlands) for valuable discussions and for her contributions to improving the quality of the manuscript. This study was supported partially by National Key R&D Program of China (2023YFA1009300, B.L.), the Science and Technology Planning Project of Guangzhou (202201020452, H.L.), and the State Key Laboratory of Mathematical Sciences, Academy of Mathematics and Systems Science, Chinese Academy of Sciences (B.L.).

**Author contributions**

B.L. conceived and initiated the study. H.L. and P.H. collected the data, preprocessed and annotated the CT scans, and compiled them into datasets. H.D., H.L., J.H., X.Y., W.L., P.C. and B.L. designed the experiments. H.D., H.L., W.C., R.L., W.P. and Y.Z. developed the code and performed the experiments. H.D. and B.L. wrote the manuscript. All authors reviewed and approved the final manuscript.

**Competing interest statement**

The authors declare no competing interests.

## Methods

### Dataset

This section describes the patient recruitment and exclusion criteria used for dataset construction.

**Patient recruitment.** The public Lung Image Database Consortium and Image Database Resource Initiative (LIDC-IDRI) dataset[13] comprises 1,018 chest CT examinations from 1,010 patients collected across seven institutions and was utilized for training, internal validation, and internal testing of deep learning models (Extended Data Fig. 1, Fig. 1e).

The single-center external test dataset was obtained from the First Affiliated Hospital of Guangzhou Medical University (National Clinical Research Centre for Respiratory Disease, Guangzhou, China) and included 913 chest CT examinations from 904 patients diagnosed with pulmonary nodules between September 2005 and July 2016.

The multi-center external test dataset comprised 443 chest CT examinations from 275 patients across 12 institutions, collected between June 2005 and August 2010, all with confirmed pulmonary nodules[14,15]. Although the multi-center dataset included cases from the First Affiliated Hospital of Guangzhou Medical University, these patients did not overlap with those in the single-center dataset.

This study was approved by the Ethics Committee of the First Affiliated Hospital of Guangzhou Medical University (approval no. 2013-43).

**Data selection.** In the LIDC dataset, each chest CT scan was independently annotated by at least three experienced thoracic radiologists, yielding 7,371 lesions labelled as pulmonary nodule by at least one reader. Lesions labelled as nodule<3 mm by at least two radiologists (n=5,984) were excluded, resulting in 1,387 nodules from 698 patients for subsequent analysis. Some patients in the filtered LIDC-IDRI dataset had multiple CT scan sequences.

For the construction of single-center and multi-center external test datasets, the following exclusion criteria were applied: (1) nodule diameter < 3 mm; (2) presence of non-solitary nodules; (3) absence of valid DICOM sequences; (4) missing diagnostic reports; (5) duplicate patient records; and (6) misdiagnosed cases. In single CT examinations of some patients, multiple slice thicknesses sequences (e.g., 1.25 mm, 2.5mm, 3.75 mm) were generated due to different data processing methods, with the moderately thick slice sequence (2.5mm) being retained.

After screening, the single-center external test dataset included 885 pulmonary nodules from 885 patients, and the multi-center external test dataset comprised 134 pulmonary nodules from 134 patients across 10 institutions. The complete data selection workflow is illustrated in Extended Data Fig. 1.

**Data preprocessing and generation of ground-truth nodule segmentation masks**

Ground-truth nodule segmentation masks for all three datasets were annotated by radiologists using a standardized workflow (Extended Data Fig. 2 and Supplementary Algorithm 1).

**Nodule mask generation for the LIDC dataset**. For CT scans in the LIDC dataset, each pulmonary nodule was independently contoured by at least three experienced thoracic radiologists. Nodule region were defined as voxels enclosed by or lying on the contoured boundaries, and only voxels identified as nodules by at least two radiologists were retained. The largest connected component of these voxels was extracted to form a three-dimensional (3D) nodule mask. To ensure spatial consistency, both the 3D nodule masks and corresponding 3D CT images were resampled to an isotropic resolution of $1 \times 1 \times 1$ mm$^3$ using nearest-neighbor interpolation for masks and cubic interpolation for CT images. CT voxel values were clipped to the range -1,000 to 1,000 Hounsfield units (HU) and linearly normalized to grayscale values (0-255). Nodule center coordinates were calculated from the resampled 3D nodule mask using the MATLAB regionprops3 function, yielding ground-truth 3D nodule masks.

**Nodule mask generation for the single-center and multi-center datasets.** For CT scans from the single-center and multi-center datasets, nodule centers were manually located by a radiologist with 12 years of experience. The original 3D CT images were resampled to a spatial resolution of $1 \times 1 \times 1$ mm$^3$ using cubic interpolation. Voxel values of the 3D CT images were clipped to -1,000 to 1,000 HU and linearly normalized to 0-255. Nodule center coordinates were proportionally scaled to maintain spatial alignment with the resampled 3D image. Based on this preprocessing, preliminary 3D nodule masks were generated using a pipeline combining deep learning-based segmentation, curve evolution, and morphological refinement, as detailed below.

First, a U-Net[16] model was employed to segment lung parenchyma, disentangling pleural-attached nodules from adjacent pleura. Second, an active contour segmentation[17] was utilized to generate an initial nodule segmentation. Third, connectivity-based filter (26-neighbor connectivity) was applied to remove isolated voxels. Morphological opening with a spherical structuring element (radius of 3 voxels) was then performed to disconnect nodule masks from adjacent bronchi or vessels, followed by morphological closing with the same structuring element to smooth irregular boundaries and fill small cavities.

All preliminary 3D nodule masks were reviewed by the radiologist, and unsatisfactory cases were manually re-segmented, resulting in the final ground-truth 3D nodule masks.

**2D mask extraction**. From the final 3D nodule masks, 2D masks of 256 × 256 pixels were extracted from the axial, coronal, and sagittal planes intersecting the resampled nodule center and used as ground-truth 2D segmentation masks.

**SAM image embeddings.** The Segment Anything Model (SAM; Fig. 1c) requires image embeddings to be computed prior to mask prediction, which is computationally intensive when using a Vision Transformer (ViT)-based image encoder[18]. To improve efficiency, image embeddings were precomputed using the ViT encoder and stored for reuse. During the human-in-the-loop interaction process of Hi-Seg (Fig. 1a), these embeddings were directly retrieved, avoiding redundant computation and enabling low-latency interactive segmentation. Embedding generation was conducted on an NVIDIA Tesla P100 GPU (16 GB memory) running Ubuntu 16.04.

## Nodule feature annotation and distribution across three datasets

This section describes the procedures used to annotate pulmonary nodule features for the three datasets and summarizes their feature distributions (Fig. 1h, i and Supplementary Table 1). Segmentation model performance was subsequently evaluated across nodule subgroups defined by these feature classifications.

**Annotation of lung nodule features in the LIDC dataset.** In the LIDC dataset, each pulmonary nodule was independently annotated by at least three thoracic radiologists. The final feature label for each nodule was determined by majority vote. Nodule size was computed as the average of the standardized binary nodule mask's long- and short-axis lengths, derived from MATLAB regionprops3 function.

Nodule density, reflecting internal texture, was graded on a five-point scale: grade 1 for ground-glass nodules, grades 2-4 for part-solid nodules, and grade 5 for solid nodules. Spiculation, indicating the presence of spike-like extensions on the nodule surface (categorized as absent, mild, or marked), was scored on a five-point scale, with a score of 1 indicating absence (label 0) and scores of 2-5 indicating presence (label 1). Lobulation, reflecting the degree of lobulated shape (absent, mild, or marked), followed the same scoring scheme as spiculation, with scores of 1 indicating absence (label 0) and 2-5 indicating presence (label 1). Calcification was scored on a six-point scale, with scores of 1-5 representing calcification patterns such as popcorn or laminated calcification (label 1) and a score of 6 indicating absence of calcification (label 0). The presence of intramodular fat was recorded as a binary variable (0, absent; 1, present).

**Annotation of lung nodule features in the single-center and multi-center testing datasets.** In both the single-center and multi-center test datasets, nodule size was defined consistently with the LIDC dataset. All other features, including density, spiculation, lobulation, calcification, cavitation, vacuole sign, pleural tag, air bronchogram, fat, and pseudo-cavity, were treated as binary variables, with 1 indicating the presence and 0 indicating its absence.

Annotations were independently performed by two senior thoracic radiologists (H.L., 12 years of experience, and P.H., 10 years of experience), both from the PET/CT Center, First Affiliated Hospital of Guangzhou Medical University. All annotations were under blinded conditions, with all other clinical information concealed. In cases of concordant assessments, the annotation was directly adopted as the final label; in cases of disagreement, a consensus label was determined through discussion.

**Lung nodule feature distribution across datasets.** The distribution of lung nodule features across the three datasets is summarized in Supplementary Table 1. Differences were observed between the LIDC dataset, which was used for model training and internal validation, and the single-center and multi-center datasets used for external testing, with most features exhibiting significantly different distributions ($P < 0.001$).

**Human participants in the human-in-the-loop collaborative segmentation**

**Participant recruitment.** We recruited five participants with varying levels of medical expertise to engage in the human-in-the-loop collaborative segmentation (Fig. 1f). The cohort comprised one senior radiologist with 12 years of experience in respiratory diseases diagnosis from the First Affiliated Hospital of Guangzhou Medical University (National Clinical Research Centre for Respiratory Diseases); one fifth-year junior medical student from Guangzhou Medical University; and three non-medical annotators from the Academy of Mathematics and Systems Science, Chinese Academy of Sciences.

**Basic radiology training for non-medical annotators.** To prepare non-medical annotators for pulmonary nodule annotation, the senior radiologist provided a 1.5-hour introductory radiology training session for the three non-medical annotators. The training covered fundamental concepts including Hounsfield unit (HU) values in CT imaging, anatomical structures and morphological features of thoracic organs on CT slices, and visual characteristics of pulmonary nodules and their corresponding HU ranges. Following training, non-medical annotators independently performed nodule localization and segmentation on randomly sampled CT images from the LIDC-IDRI dataset. Their annotations were subsequently reviewed by the senior radiologist, who provided feedback on segmentation accuracy and boundary delineation.

**Hi-Seg operation training.** All five participants underwent a standardized training session to familiarize themselves with Hi-Seg using CT examinations from 10 patients in the LIDC-IDRI dataset (30 two-dimensional slices in total). Participants practiced interactive segmentation through the Hi-Seg user interface (Fig. 1g), iteratively adjusting prompt placements to obtain high-quality segmentation with fewer interactions. The training session lasted 1.5 hours.

**Formal segmentation experiments using Hi-Seg.** To evaluate segmentation performance under the human-in-the-loop paradigm, each participant performed pulmonary nodule segmentation using Hi-Seg and, separately, independent manual annotation. Both approaches were applied to identical single-center and multi-center external testing datasets. To minimize recall and learning effects, a four-week washout period was imposed between the two segmentation conditions.

**User interface for manual and human-in-the-loop segmentation based on SAM**

We developed a user interface (UI) for chest CT image segmentation that supports two operational modes: manual segmentation and human-in-the-loop segmentation based on SAM, termed Hi-Seg (Fig. 1a). The UI was implemented in Python 3.9 utilizing PyTorch (v1.13.1), torchvision (v0.14.1), OpenCV (v4.7.0), Segment Anything, and PyQt6.

**Interface layout and functionality.** The UI consists of two main panels.

Left panel. The top section displays patient metadata, including dataset name, case identifier, and the current 2D CT view (axial, coronal, or sagittal). The bottom section contains functional buttons:

- Reset: In Hi-Seg mode, clears all prompt points; in manual mode, removes all delineated nodule boundaries.
- Undo: In Hi-Seg mode, deletes the most recently added prompt point and reverts the segmentation mask to its previous state; in manual mode, removes the most recently drawn nodule boundary and its associated mask.
- Cut-Outs: Finalizes the segmentation for the current CT image. In Hi-Seg mode, the final mask and the annotator’s prompt-point trajectory are saved, and the next image is automatically loaded; in manual mode, the current mask is saved and the next image is loaded.

Right Panel. Displays the current 2D chest CT image, where annotators perform segmentation in either Hi-Seg or manual mode.

**Segmentation modes.** The UI supports both manual segmentation and Hi-Seg modes.

To ensure experimental isolation, the two segmentation modes were implemented as separate executable programs.

Hi-Seg mode. In the human-in-the-loop setting, annotators iteratively place prompt points on the CT image, and SAM generates segmentation masks in response:

- Left click: Adds a blue prompt point to include regions in the segmentation mask.
- Right click: Adds a yellow prompt point to exclude undesired regions from the mask.

Manual segmentation mode. Annotators manually delineate nodule contours by holding the mouse button. Upon release, the start and end points are automatically connected to form a closed curve, which is filled to generate the segmentation mask.

- Hold left button: Draws a blue curve to include pixels within the region of interest.
- Hold right button: Draws a yellow curve to exclude pixels from the region of interest.

**Image display and coordinate handling.** To improve visual clarity and operational precision, CT images are upscaled fourfold to a display resolution of 1,024×1,024 pixels using bicubic interpolation. Segmentation masks are upscaled by the same factor using nearest-neighbor interpolation. Prompt-point coordinates placed by annotators on the upscaled image are divided by four before being input into SAM, ensuring accurate spatial alignment between user input and model inference.

**Segment Anything Model (SAM): architecture, training, variants, and hyperparameter settings**

Segment Anything Model (SAM) is a foundation segmentation model pretrained on large-scale datasets that produces object masks conditioned on prompts, such as points or bounding boxes, enabling prompt-driven and interactive image segmentation.

**SAM architecture.** SAM[4] comprises three core components: an image encoder, a prompt encoder and a mask decoder (Fig. 1c). The image encoder is a Masked Autoencoder (MAE)[49]-pretrained Vision Transformer (ViT)[18] with 32 transformer layers, which maps the input image into a high-dimensional latent representation. The prompt encoder converts prompts (points or bounding boxes) into positional encodings using linear and trigonometric transformations, guiding the model toward target regions. Multiple prompts can be encoded simultaneously, with their embeddings concatenated to enable progressive refinement of masks through iterative prompt addition. When a coarse mask is provided as a prompt, it is encoded using three convolutional layers and incorporated as an additional prompt signal to refine the segmentation. The mask decoder consists of two transformer layers, two transposed convolutional layers and four two-layer multilayer perceptrons (MLP#1-#4), which integrate image and prompt embeddings to produce a continuous-valued segmentation mask. Each pixel value represents the predicted probability of belonging to the target region. These probability maps are binarized using a predefined threshold to generate the final segmentation mask, together with a confidence score reflecting model's internal estimate of mask quality.

**SAM training.** SAM was trained on the SA-1B dataset, which comprises 11 million natural images annotated with over 1 billion masks. The dataset spans a broad range of objects and scenes, from large object masks to fine-grained delineations. During training, point and box prompts were randomly generated using predefined algorithms[50,51], analogous to the pseudo-human prompting strategies employed in our study. Each mask undergoes up to 11 iterative refinement rounds to simulate interactive segmentation.

**Prompt-free SAM.** Prompt-free SAM, also referred to as SAM everything mode[4], enables end-to-end segmentation of all identifiable objects in an input image without human intervention. A uniform 32×32 grid of sampling points is generated over the image, with each point treated as an independent prompt and producing three candidate masks. Masks are filtered in batches of 64, where masks with low confidence scores or insufficient stability under varying thresholds are discarded. Among the remaining masks, the following masks are removed: (i) masks that overlap with others but exhibiting lower confidence scores, (ii) low-resolution masks and (iii) masks with small spatial extent. The surviving masks constitute the final output, each corresponding to a distinct object instance in the image.

**SAM variants.** We evaluated three SAM variants: HQ-SAM[24], MedSAM[7] and MedSAM2[25]. These variants differ through structural modifications or fine-tuning on specialized datasets. HQ-SAM incorporates (i) global-local feature fusion within the image encoder, (ii) high-quality output tokens injected into the mask decoder and (iii) fine-tuning on the HQSeg-44K dataset, which contains 44,320 high-quality annotated masks. MedSAM employs a ViT-B backbone as the image encoder and is fine-tuned on a large-scale medical imaging dataset comprising 1,570,263 images across 10 modalities and 30 cancer types. As MedSAM is fine-tuned exclusively using box prompts, only box-prompted MedSAM was evaluated in our study. MedSAM2 builds upon SAM2[26], which replaces the original image encoder with a hierarchical attention network, augments the mask decoder with an additional head and skip connections, and incorporates a memory to propagate cross-frame information for video inputs. Building on this, MedSAM2 further incorporates a self-sorting memory bank to better accommodate unordered medical images, and is fine-tuned on the One-Prompt dataset, which integrates 78 public medical datasets spanning over 10 modalities and more than 3,000 clinician-annotated samples. To ensure experimental fairness, MedSAM2 was applied to 2D images only.

**SAM hyperparameters.** Two hyperparameters, mask_input and multimask_output significantly influence segmentation behavior. The mask_input parameter controls whether the predicted mask from the current interaction is fed back into the prompt encoder alongside new prompts to guide subsequent refinements. The multimask_output parameter controls whether multiple candidate masks are generated for ambiguous prompts: when enabled, three masks decoded by MLP #2-#4 are produced, and the one with the highest confidence score is selected; when disabled, a single mask decoded by MLP #1 is returned. In our human-in-the-loop experiments, annotators preferred enabling mask_input while disabling multimask_output, yielding more stable and controllable outcomes. These hyperparameter settings were therefore applied across all SAM-based models to ensure fair comparison.

## Pseudo-human prompts

Pseudo-human prompts were algorithmically generated from ground-truth masks to mimic human point- or box-prompting behaviours (Fig. 1c), a common surrogate for real human annotators in SAM-based segmentation studies[4-6]. Five representative pseudo-human prompting strategies were evaluated: four point-based prompts (pseudo-human prompts #1-#4) and one bounding-box prompt.

Pseudo-human prompts #1[5] and #2[4] are iterative strategies that generate new points based on the model's current segmentation. At each iteration, the predicted mask (initialized as empty) is compared with the ground-truth mask using an exclusive-or (XOR) operation to identify erroneous regions, from which new prompt points are sampled. The two strategies differ only in how new points are generated. Pseudo-human prompt #1 selects the point farthest from the error-region boundary, with a random 0-5 pixel perturbation while remaining within the target region. In contrast, pseudo-human prompt #2 randomly samples a point from the erroneous region; points are labelled as left clicks if they fall inside the ground-truth mask and as right clicks otherwise. All points generated across iterations are cumulatively provided to the model together with the latest mask prediction until a predefined number of prompts is reached.

Pseudo-human prompts #3 and #4 are non-iterative strategies in which all prompt points are generated from the ground-truth mask before model inference[6]. Pseudo-human prompt #3 randomly generates five points within the target region. Pseudo-human prompt #4 generates five foreground points within the target region and five background points within an expanded bounding box centered on the target region, with side lengths three times those of the tight bounding box. All points are provided simultaneously to SAM, yielding a single segmentation mask.

Bounding-box prompt generates the smallest bounding box that fully encloses the ground-truth mask.

SAM, HQ-SAM, and MedSAM2 were evaluated using pseudo-human prompts #1-#4, whereas MedSAM was evaluated using bounding-box prompts, consistent with its original design. For fairness, pseudo-human prompts #1 and #2 used the same number of points per image as human annotators, defined by the median number of clicks across five annotators. The median number of points per image was 2.27 for the single-center dataset and 2.31 for the multi-center dataset. Pseudo-human prompts #3 and #4 used 5 and 10 points per image, respectively.

**Training of deep learning models for lung nodule segmentation**

We trained five state-of-the-art deep learning models for lung nodule segmentation, namely U-Net[19], U-Net++[20], AttentionU-Net[21], TransU-Net[22] and nnU-Net[8]. Model training and internal validation were performed using the publicly available LIDC-IDRI chest CT dataset[13], which was randomly split into training and validation sets at an 8:2 ratio (Fig. 1e, Extended Data Fig. 1and Extended Data Fig. 2).

For all models, the training objective was defined as the sum of binary cross-entropy (BCE) loss and Dice loss:

$$L = L_{\text{BCE}} + L_{\text{Dice}}. \quad (1)$$

The BCE loss quantifies the discrepancy between the predicted pixel-wise class probabilities and ground-truth labels and is defined as:

$$L_{\text{BCE}} = -\frac{1}{N}\sum_{i=1}^{N}[g_i \log s_i + (1 - g_i)\log(1 - s_i)], \quad (2)$$

where $N$ is the total number of pixels in the image, $g_i \in \{0,1\}$ is the ground-truth label for the $i$-th pixel, and $s_i$ is the corresponding predicted probability.

The Dice loss evaluates the spatial overlap between predicted and ground-truth segmentation masks and is defined as:

$$L_{\text{Dice}} = 1 - \frac{2\sum_{i=1}^{N} g_i s_i}{\sum_{i=1}^{N} g_i^2 + \sum_{i=1}^{N} s_i^2}. \quad (3)$$

During training, nnU-Net adaptively optimized its training strategy based on dataset characteristics. For U-Net, U-Net++, Attention U-Net, and TransU-Net, optimization was performed using the Adam optimizer with an initial learning rate of 1e-5 and a batch size of 32. All models were trained for 300 epochs. The model checkpoint achieving the highest Dice score on the internal validation set was selected for subsequent external evaluation.

## Performance metrics

We employed three metrics to evaluate segmentation accuracy, segmentation efficiency, and failure rate.

**Segmentation accuracy.** Segmentation accuracy was quantified using the Dice similarity coefficient (Dice score)[6,7], defined as

$$\text{Dice}(G, S) = \frac{2|G \cap S|}{|G| + |S|}, \tag{4}$$

where $G$ denotes the ground-truth mask, $S$ represents the predicted segmentation mask, and $|\cdot|$ indicates the number of pixels within the mask. This metric quantifies the spatial overlap between predicted and ground-truth segmentations.

**Segmentation efficiency.** Segmentation efficiency was measured in seconds as the cumulative time required to complete nodule segmentation across the axial, coronal, and sagittal planes. Timing started when the CT image was displayed in the user interface and ended when the annotator clicked the Cut-Outs button. Both manual and Hi-Seg-assisted segmentation times were recorded. To mitigate the influence of outliers caused by delayed interface closure, any single segmentation session exceeding 300s was excluded. Furthermore, if the segmentation time in any of the three planes exceeded 300s, the cumulative time for that nodule was discarded to ensure the reliability of efficiency measurements.

**Failure rate.** Failure rate was defined as[52]

$$\text{Failure rate} = \frac{1}{M} \sum_{i=1}^{M} \mathbb{I}(\text{Dice}(G_i, S_i) \leq 50\%), \tag{5}$$

where $M$ is the total number of images, $G_i$ is the $i$-th ground-truth mask, $S_i$ is the corresponding predicted segmentation mask, $\text{Dice}(\cdot,\cdot)$ is the Dice similarity coefficient, and $\mathbb{I}(\text{condition})$ is the indicator function, equal to 1 if the condition is satisfied and 0 otherwise.

## Statistical analysis

Continuous variables (for example, nodule sizes) were reported as medians with interquartile ranges (25$^{th}$ and 75$^{th}$ percentiles). Categorical variables, including nodule characteristics (for example, density and spiculation), were represented as counts and percentages. Dice scores and segmentation time were presented as means with 95% confidence intervals. Failure rates were represented as percentages.

Differences in nodule characteristics across datasets were evaluated using the Kruskal-Wallis *H* test for continuous variables and Chi-square tests with Bonferroni corrections for categorical variables. Comparisons of Dice scores between segmentation models and of segmentation time between manual annotation and Hi-Seg-assisted segmentation were performed using the Wilcoxon signed-rank test. Model performance across nodule subgroups was assessed using the Friedman tests, followed by Dunn & Šidák's post-hoc multiple comparison test.

All statistical tests are two-sided, and $P<0.05$ was considered statistically significant. Statistical analyses were conducted using MATLAB R2020a (MathWorks).

**Reporting summary**

Further information on research design is available in the Nature Portfolio Reporting Summary linked to this article.

**Ethics approval and consent to participate**

This study was approved by the Ethics Committee of the First Affiliated Hospital of Guangzhou Medical University, Guangzhou, China (approval no. 2013-43), and was conducted in accordance with the Declaration of Helsinki. The study utilized the publicly available LIDC-IDRI chest CT dataset. As this dataset is fully anonymized and publicly accessible, additional institutional review board approval and individual patient consent were not required. In addition, retrospective single-center and multi-center CT imaging data were used with approval from the institutional ethics committee and in compliance with all applicable ethical and regulatory requirements. The requirement for informed consent was waived due to the retrospective nature of the study and the use of anonymized imaging data. Five human participants with varying levels of medical expertise were recruited to participate in the human-in-the-loop collaborative segmentation experiments. All participants provided informed consent prior to participation.

**Data availability**

The LIDC-IDRI imaging dataset is publicly available at https://www.cancerimagingarchive.net/collection/lidc-idri/. Single-center and multi-

center imaging data were collected with approval from the institutional ethics committee and in accordance with all applicable ethical and regulatory requirements of our institution. De-identified tabular data generated in this study are available from the corresponding author upon reasonable request for non-commercial research purposes. Requests should be directed to H.P.L. (lhp586@163.com). A data use agreement may be required.

## Code availability

The code is available on GitHub at https://github.com/DDDHQ/HI-Seg.

## Methods-only references

# Figures

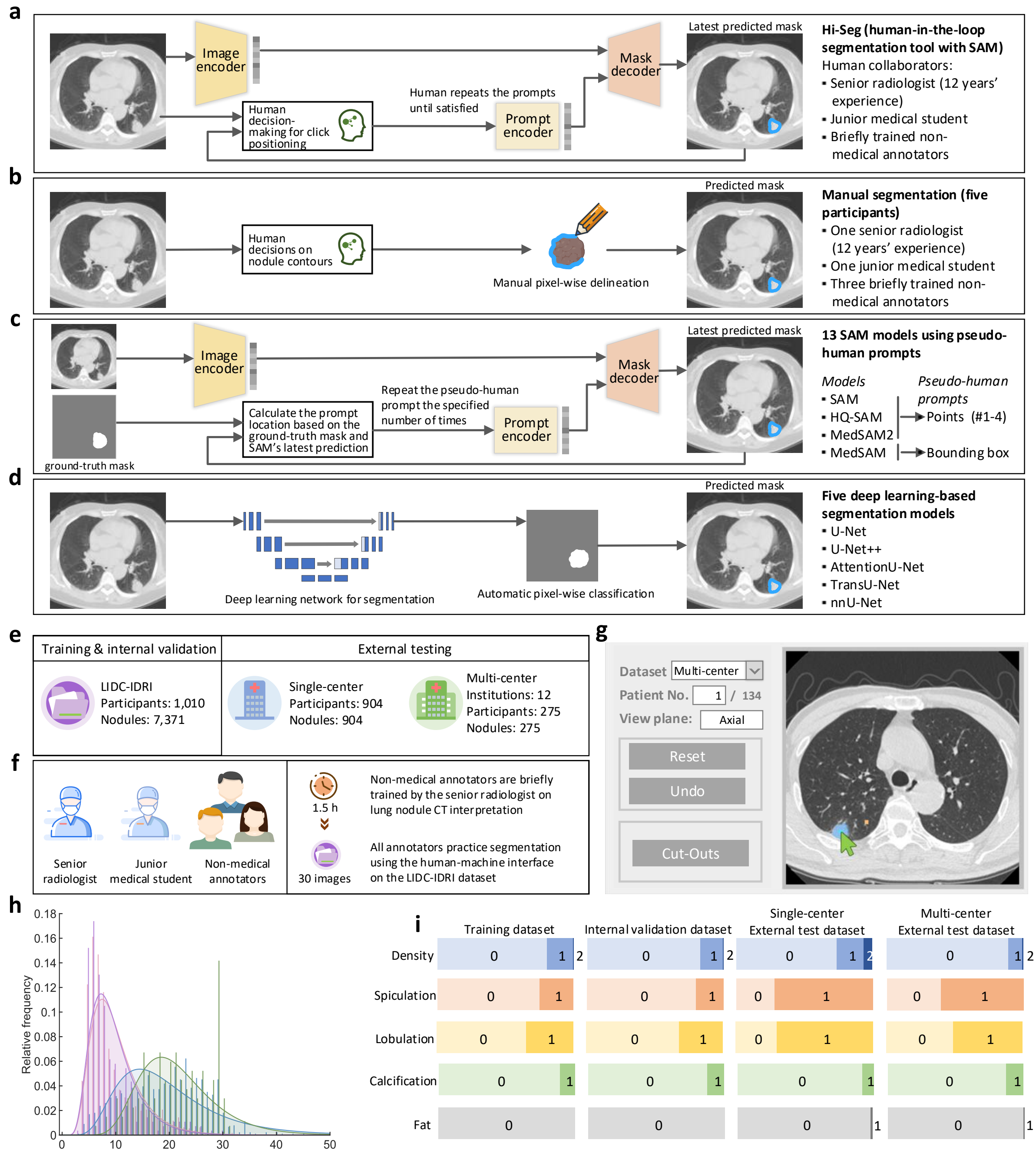

## Fig. 1: Overview of the study design.

**a-d,** Comparison of Hi-Seg with state-of-the-art segmentation approaches. **a,** Hi-Seg, a human-in-the-loop segmentation framework for pulmonary nodules built on Segment Anything Model (SAM). Humans iteratively refine prompts through trial-and-error learning and semantic reasoning, progressively guiding SAM toward higher-quality masks.

**b,** In manual segmentation, annotators draw curves along the perceived nodule contours; pixels enclosed by the curves constitute the segmentation masks.

**c,** Pseudo-human prompts are automatically generated from ground-truth masks to simulate real human annotators; prompt are repeated a predefined number of times.

**d,** Deep learning models automatically classify whether each pixel belongs to the target nodule.

**e**, The LIDC-IDRI dataset was used for training and internal validation of the deep learning models, whereas single-center and multi-center datasets were used for external testing.

**f,** Annotators with and without medical backgrounds were recruited for human-machine collaboration. All annotators received training in prompt-based segmentation using Hi-Seg, and those without medical backgrounds received additional brief training in chest CT interpretation.

**g**, The Hi-Seg user interface, illustrating the framework for human-SAM interaction in medical image segmentation.

**h**, Distribution of nodule sizes, overlaid with corresponding Gaussian fitted curves.

**i**, Distribution of nodule density, spiculation, lobulation, calcification and fat. Nodule density was categorized as follows: 0, solid nodules; 1, part-solid nodules; and 2, ground-glass nodules (GGNs). For spiculation, lobulation, calcification and fat, a value of 1 indicates presence of the characteristic, whereas 0 indicates absence.

Hi-Seg refers to the proposed Human-in-the-loop segmentation framework that integrates real-time interactive prompts provided by human annotators. Segment Anything Model (SAM) is a prompt-driven foundation model that generates segmentation masks from user-provided prompts, such as points or bounding boxes. Deep-learning models include U-Net, U-Net++, Attention U-Net, TransUNet and nnU-Net.

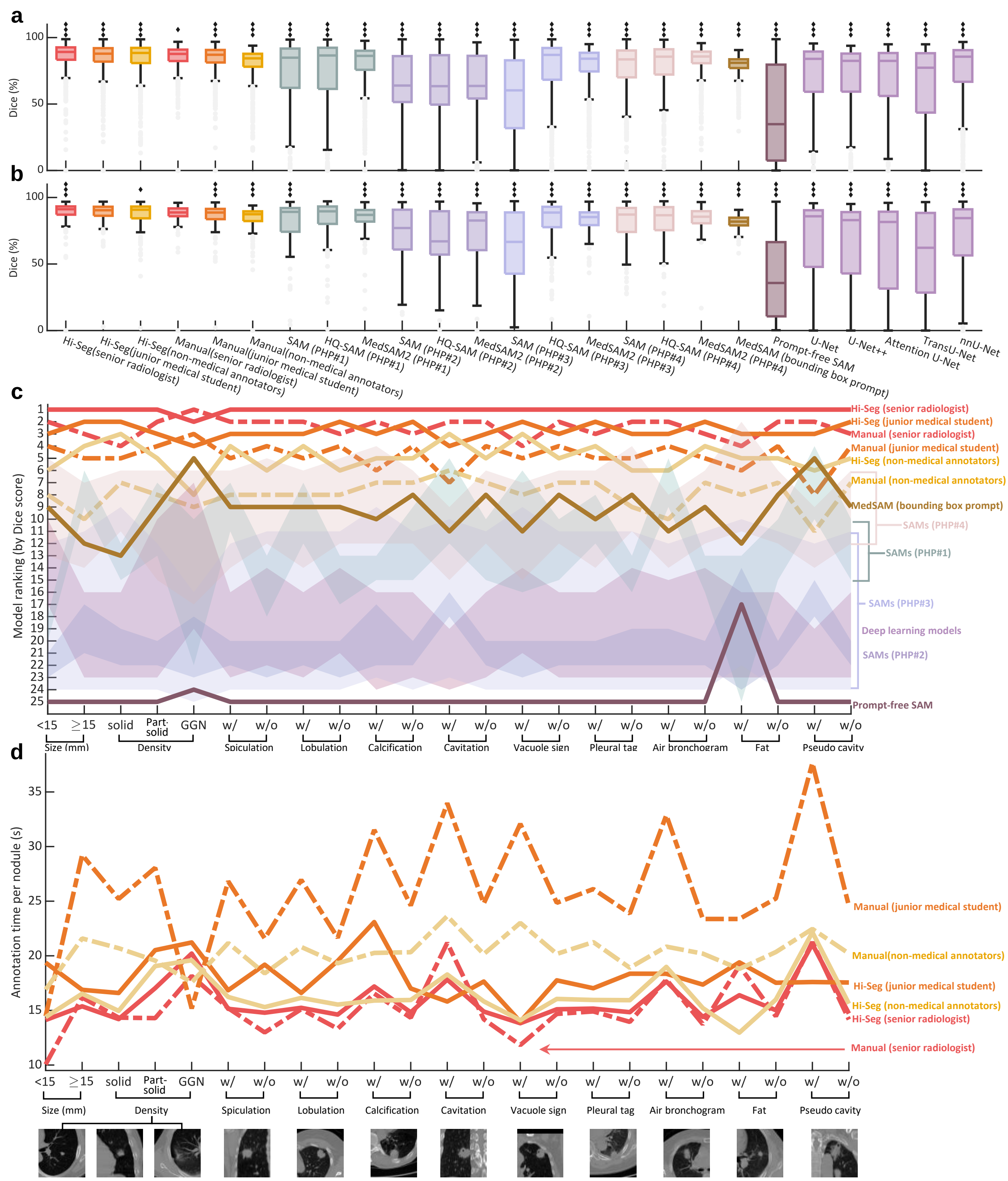

**Fig. 2: Model performance evaluation.**

**a,b,** Segmentation accuracy (Dice score) of each model on the single-center (a) and multi-center (b) external test datasets. Box plots show the median (center line), interquartile range (25th-75th percentiles), and outliers (light grey points). Black dots above boxes indicate statistical significance relative to the mean performance of five annotators using Hi-Seg: one dot denotes $P < 0.05$, and three dots denote $P < 0.001$.

**c,** Segmentation accuracy rankings of each model stratified by pulmonary nodule subgroup. For each subgroup ("w/" and "w/o" indicate the presence or absence of the characteristic), models were ranked based on the mean Dice scores of nodules from that subgroup across two external test datasets.

**d,** Annotation time per nodule (seconds) of each model for each pulmonary nodule subgroup. Annotation time was defined as the cumulative segmentation time across axial, coronal and sagittal planes and averaged within each subgroup.

Hi-Seg and Manual (senior radiologist/junior medical student) denote segmentation by the corresponding annotator using Hi-Seg or manual delineation, respectively; Hi-Seg and Manual (non-medical annotators) denote mean performance across three non-medical annotators using Hi-Seg or manual delineation, respectively. For SAM-based models (SAM, HQ-SAM, MedSAM2 and MedSAM), the prompts indicated in parentheses denote the pseudo-human prompt used (for example, HQ-SAM (PHP#1)). Prompt-free SAM generates masks in a single pass without prompts. Shaded regions labelled as SAMs (PHP indices) in (c) indicate ranking ranges achieved by SAM, HQ-SAM and MedSAM2 under the corresponding pseudo-human prompting strategy; shaded regions for deep learning models in (c) indicate ranking ranges across U-Net, U-Net++, Attention U-Net, TransU-Net and nnU-Net.

Hi-Seg, proposed human-in-the-loop segmentation framework; SAM, Segment Anything Model; HQ-SAM, MedSAM2 and MedSAM, SAM variants; PHP, pseudo-human prompt; GGN, ground-glass nodule; U-Net, U-Net++, Attention U-Net, TransU-Net and nnU-Net, deep learning architectures.

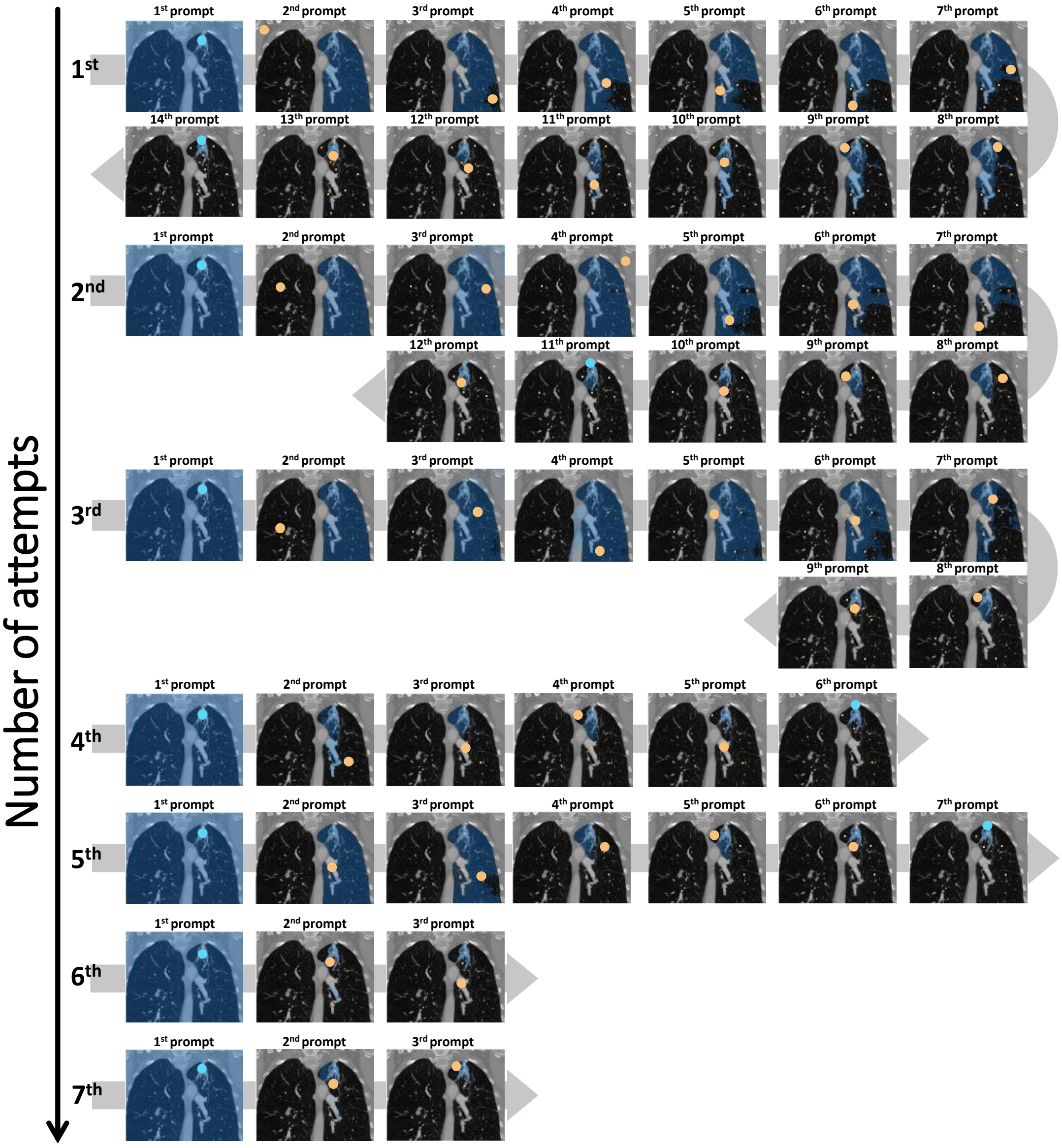

**Fig. 3: Iterative feedback refinement accelerates mask convergence toward the ground truth.** A senior radiologist with 12 years of experience in respiratory diseases performed seven independent segmentation attempts on the same CT image, assisted by Hi-Seg. The attempts are shown sequentially from top to bottom. In each attempt, the predicted mask was iteratively updated in response to radiologist-provided prompts, with refinement directions indicated by gray arrows. Blue prompt points add regions to the mask, yellow points remove redundant areas, and light-blue overlays show intermediate masks generated after each prompt. The most recent prompt point is enlarged; earlier prompts retain their original size. In the initial attempt, the Hi-Seg-generated mask overestimated the target nodule, nearly covering the entire image. The radiologist subsequently explored alternative prompt placement strategies to refine mask boundaries. During the first three attempts, second prompts placed far from the nodule removed only limited irrelevant regions. From the fourth attempt onward, prompts were positioned progressively closer to the nodule boundary, effectively reducing extraneous regions. By the sixth and seventh attempts, accurate segmentation was achieved with only three prompts, compared with 14 prompts in the first attempt.

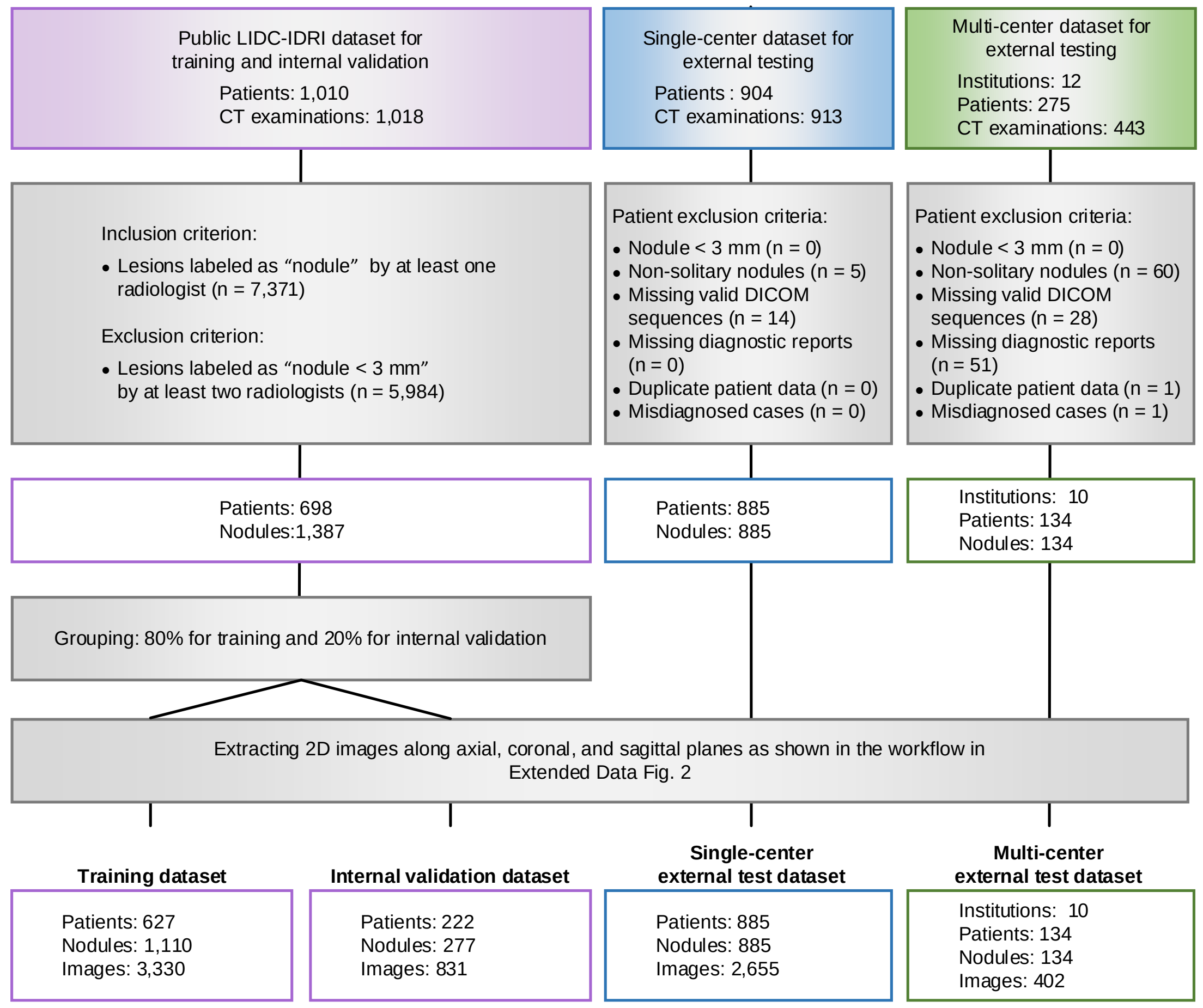

**Extended Data Fig. 1: Preprocessing and filtering of the datasets. Diagram describing the inclusion and exclusion in this study.**

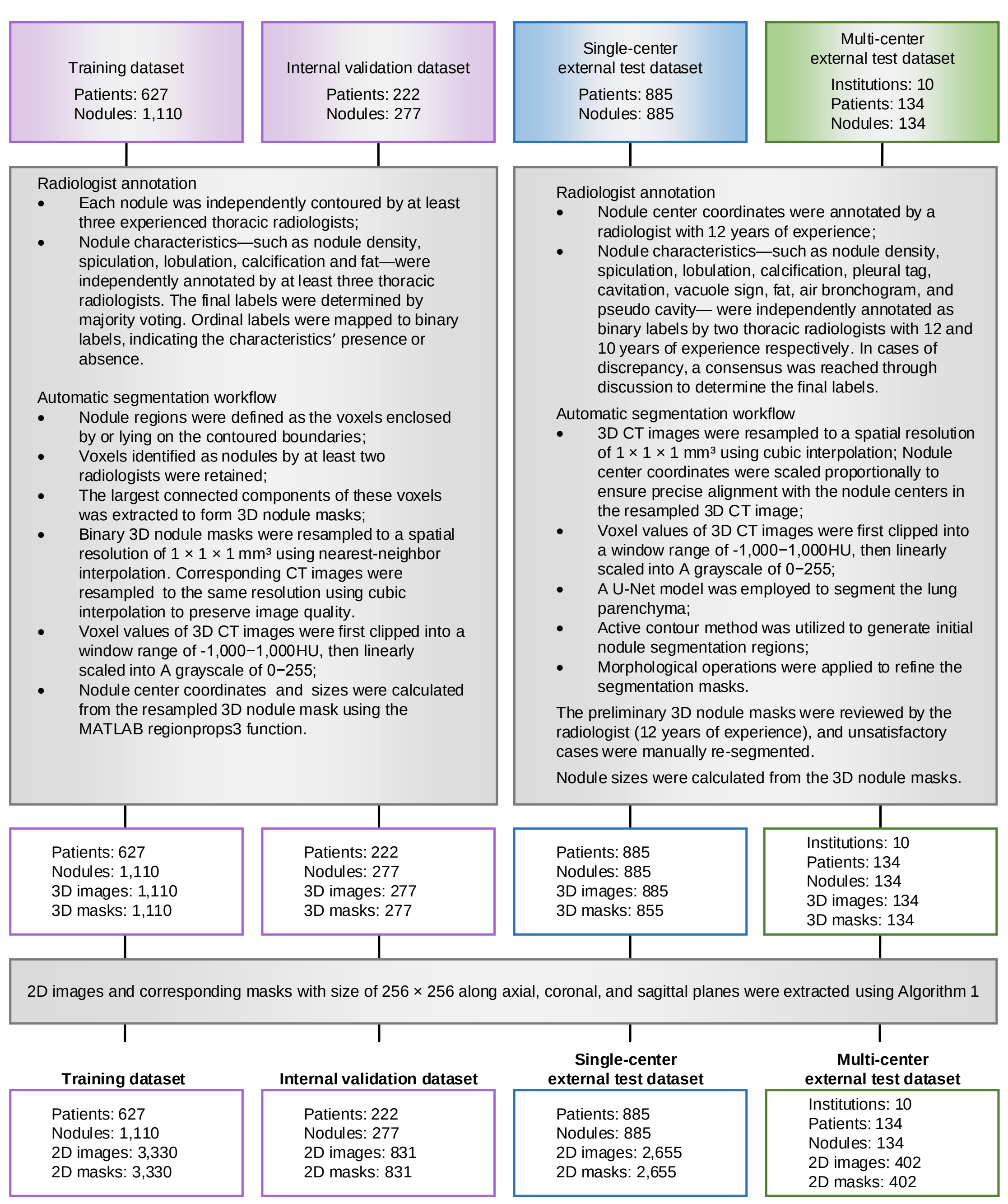

**Extended Data Fig. 2: Data preprocessing and generation of ground-truth nodule segmentation masks.**

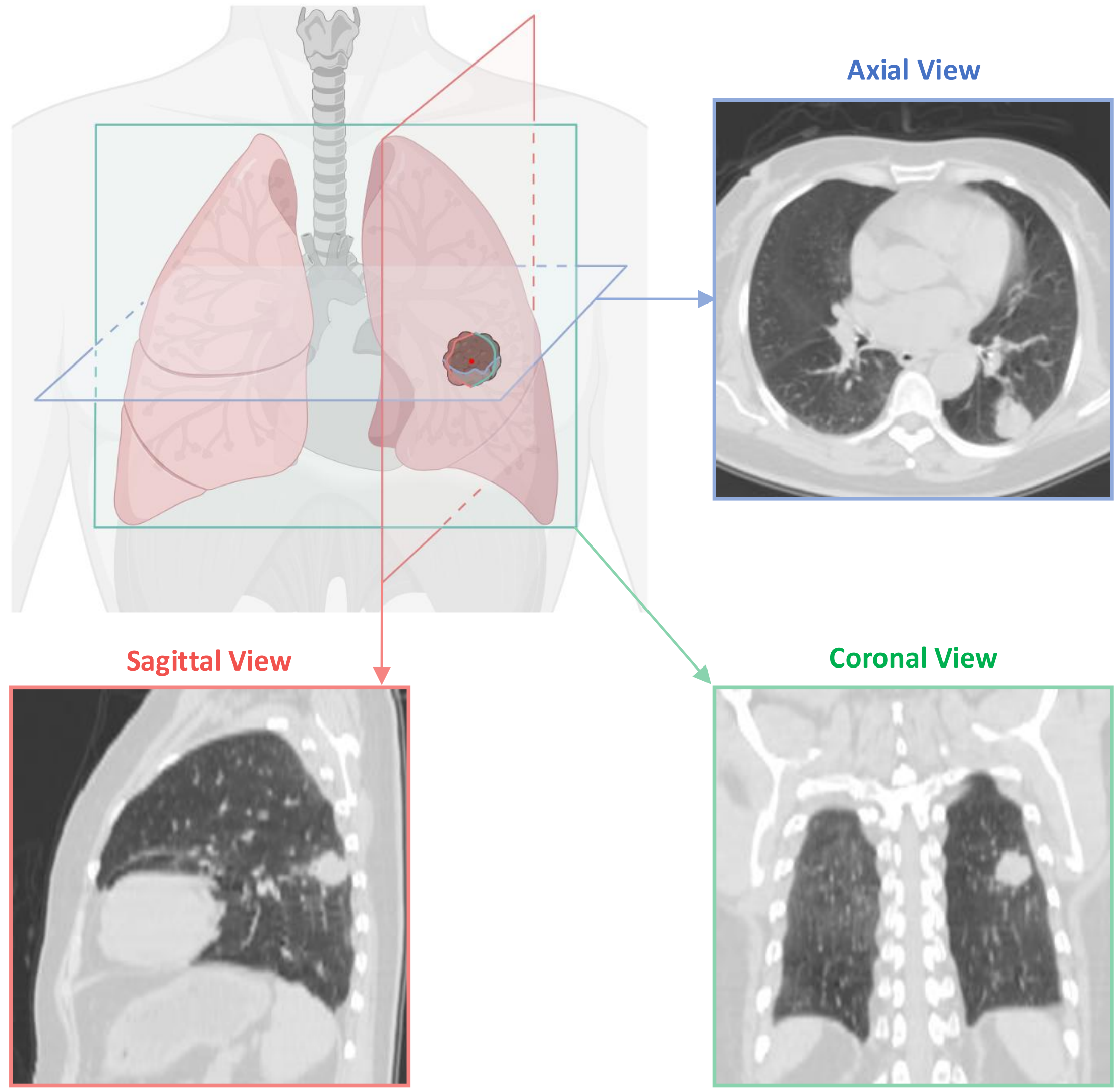

**Extended Data Fig. 3: Human-in-the-loop interactive segmentation across axial, coronal and sagittal CT views.** Annotators sequentially localize and segment pulmonary nodules from axial, coronal, and sagittal two-dimensional slices of chest CT images.

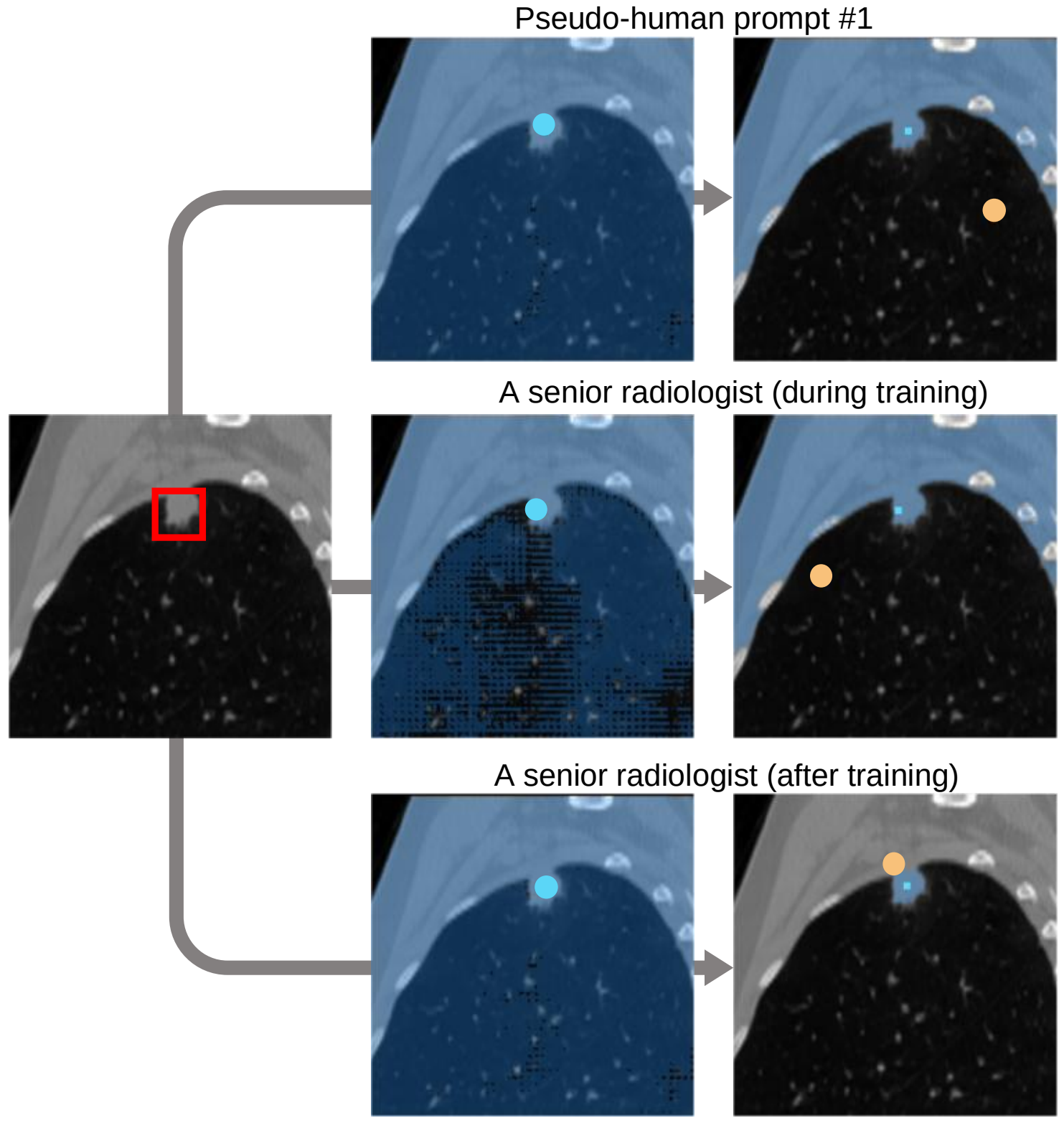

**Extended Data Fig. 4: Effects of prompt placement across three prompting strategies on predicted masks for pleural adhesion nodules.** Pleural adhesion nodules lie within the lung parenchyma and closely abut on the pleura. CT images show predicted masks generated using the best-performing iterative pseudo-human strategy (Pseudo-human prompt #1), a senior radiologist during training, and the same radiologist after training. Note that the best-performing pseudo-human prompt (#4) is non-iterative that generates all points at once, and thus cannot be compared to human annotators. Columns display the cropped CT image (left), the first prompt and resulting mask (middle), and the second prompt and resulting mask (right). Blue points add mask regions, yellow points remove redundant regions, and light blue indicates predicted masks. Current prompt points are enlarged, whereas previous points retain their original size. For all strategies, the first prompt is placed at the nodule center, yielding oversized masks. For the second prompt, the pseudo-human strategy and the radiologist during training place prompts within the lung parenchyma, partially removing redundant regions but leaving residual masks in adjacent structures. In contrast, after training, the radiologist places the prompt at the nodule-pleura interface, eliminating all redundant regions in a single step.

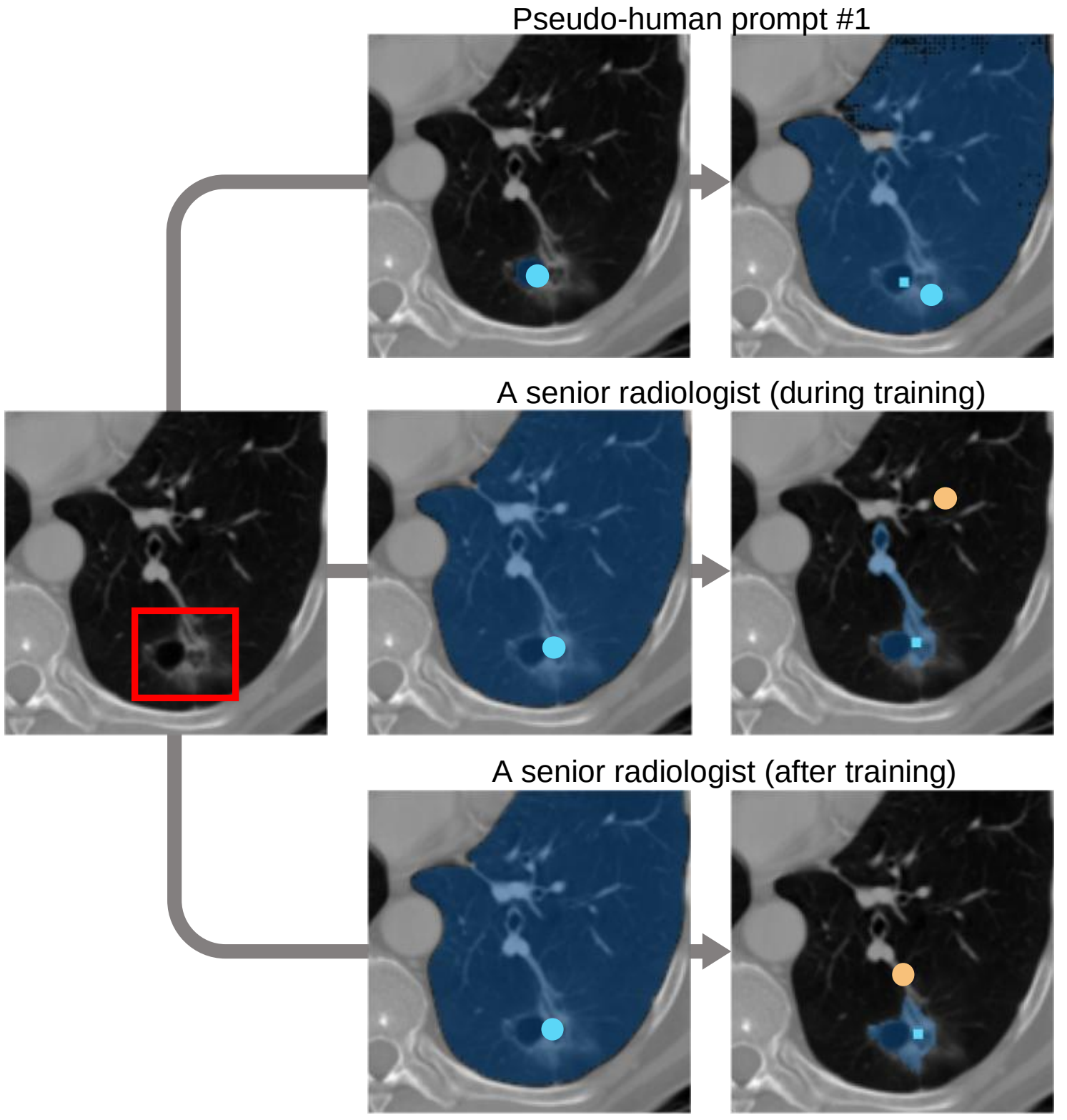

**Extended Data Fig. 5: Effects of prompt placement across three prompting strategies on predicted masks for cavitary nodules.**

Cavitary nodules are characterized by a high-density wall surrounding a low-density cavity. Under the pseudo-human strategy, the first prompt placed within the cavity segments only the cavity and misses the nodule wall; the second prompt placed on the wall produces an oversized mask that includes surrounding lung parenchyma. The radiologist during training places the first prompt on the nodule wall, similarly yielding an oversized mask, and the second prompt in distant parenchyma partially removes non-target regions but fails to exclude attached bronchi. After training, the radiologist targets the attached bronchi with the second prompt, successfully eliminating all redundant regions.

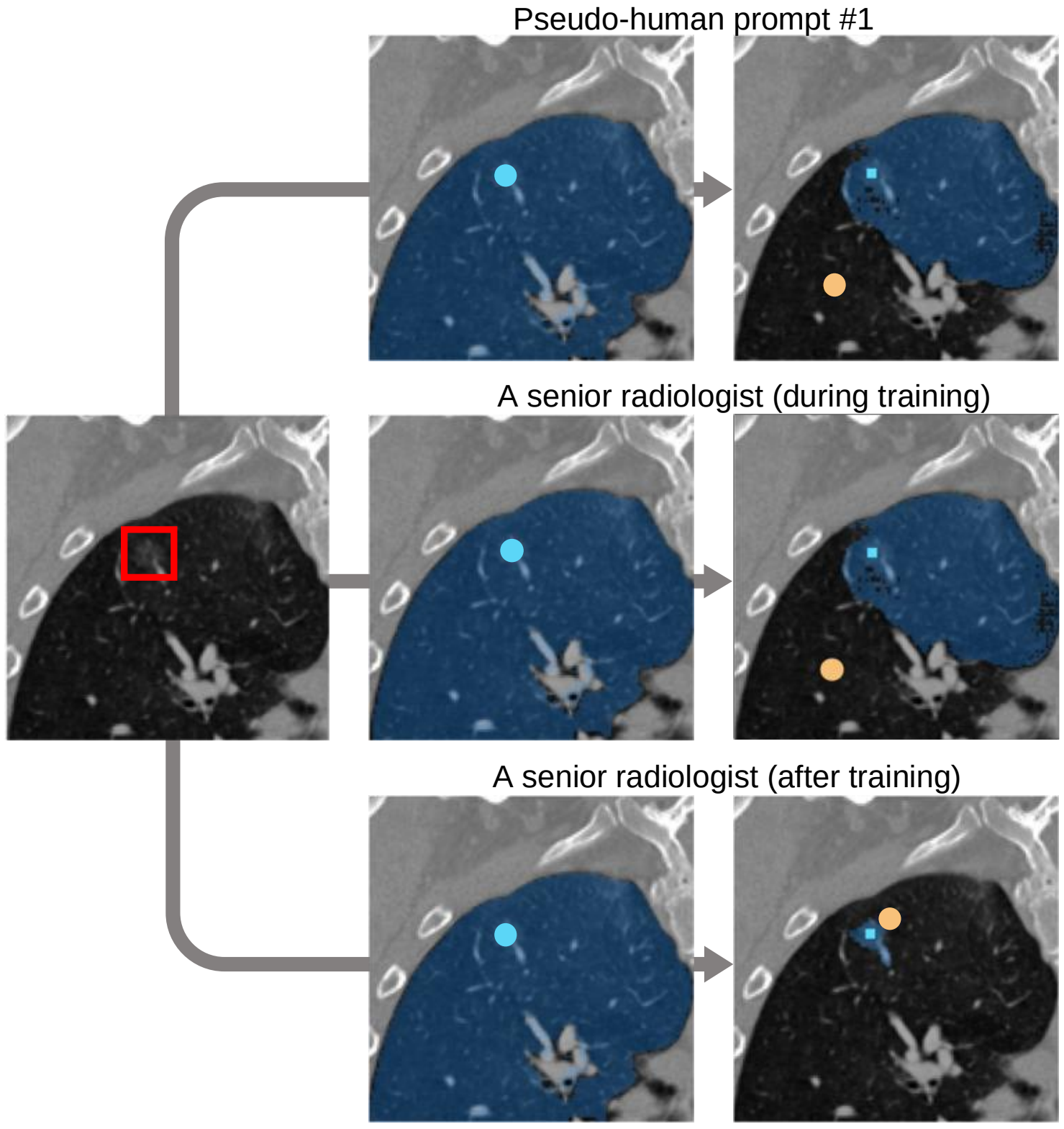

**Extended Data Fig. 6: Effects of prompt placement across three prompting strategies on predicted masks for ground-glass nodules.**

Ground-glass nodules characterized by subtle haze-like attenuation with texture resembling normal parenchyma. All strategies place the initial prompt at the nodule center, resulting in oversized masks due to indistinct boundary. The placement of the second prompt differs across strategies and leads to divergent outcomes. The pseudo-human prompt #1 places the second prompt in distant parenchyma, achieving only limited mask reduction. The radiologist during training places the prompt on attached tissue, failing to delineate the target. After training, the radiologist avoids prompting within the ground-glass region or attached tissue and instead places a background-removal prompt in adjacent parenchyma, leading to improved boundary delineation.

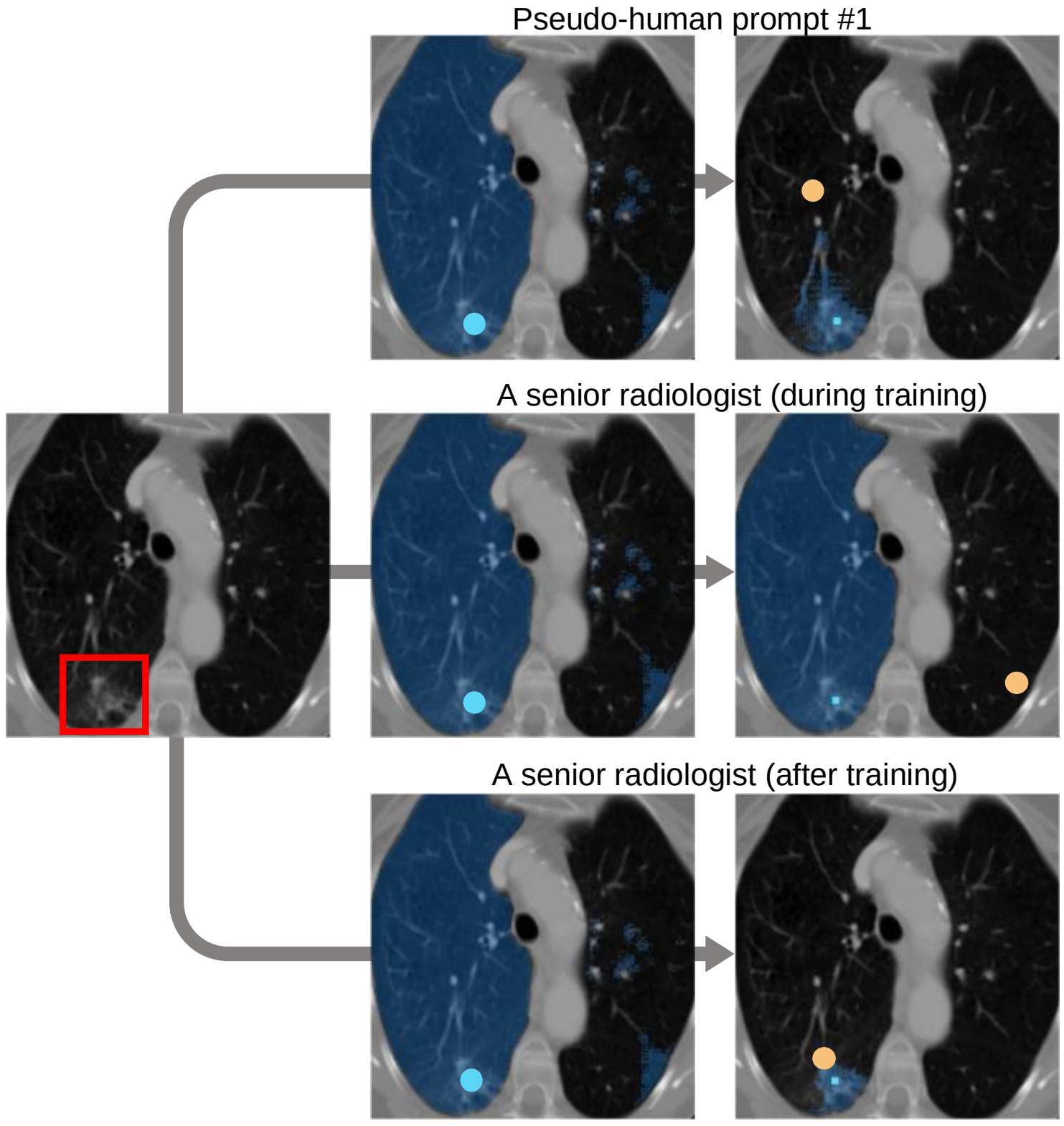

**Extended Data Fig. 7: Effects of prompt placement across three prompting strategies on predicted masks for pleural traction nodules**.

Pleural traction nodules are characterized by pleural indentation toward the nodule. Initial center-point prompts include ipsilateral parenchyma and generate spurious masks in the contralateral lung. With the second prompt, the pseudo-human prompt #1 targets distant parenchyma, yielding limited correction, whereas the radiologist during training removes contralateral noise. After training, the radiologist prioritizes the dominant erroneous region adjacent to the nodule, eliminating both ipsilateral and contralateral noise in a single interaction.

## Tables

Table 1. Comparison of segmentation performance between Hi-Seg with real human prompts, deep learning models, and Segment Anything-based methods using pseudo-human prompts across two external test datasets.

| Model | Single-center (*n*=885) | | Multi-center (*n*=134) | |
|---|---|---|---|---|
| | Dice (%)[a] | *P*-value[b] | Dice (%)[a] | *P*-value[b] |
| **Deep learning models[c]** | | | | |
| U-Net | 70.77 (68.99-72.56) | < 0.001 | 66.52 (61.02-72.01) | < 0.001 |
| U-Net++ | 69.84 (68.07-71.61) | < 0.001 | 65.25 (59.88-70.62) | < 0.001 |
| Attention U-Net | 67.88 (65.96-69.80) | < 0.001 | 63.25 (57.61-68.88) | < 0.001 |
| TransU-Net | 63.51 (61.54-65.48) | < 0.001 | 56.89 (51.08-62.70) | < 0.001 |
| nnU-Net | 75.42 (73.90-76.94) | < 0.001 | 69.05 (64.21-73.88) | < 0.001 |
| **Segment anything models / Pseudo-human prompts[d]** | | | | |
| SAM (overall, prompts #1-4) | 67.77 (66.93-68.62) | < 0.001 | 73.85 (71.93-75.77) | < 0.001 |
| HQ-SAM (overall, prompts #1-4) | 73.49 (72.73-74.26) | < 0.001 | 78.11 (76.45-79.78) | < 0.001 |
| MedSAM2 (overall, prompts #1-4) | 76.12 (75.49-76.76) | < 0.001 | 80.90 (79.70-82.11) | < 0.001 |
| MedSAM (bounding box prompt) | 79.24 (78.76-79.72) | < 0.001 | 81.50 (80.60-82.40) | < 0.001 |
| **Human-in-the-loop segmentation (Hi-Seg)[e]** | | | | |
| Hi-Seg (overall, five human annotators) | 84.47 (84.05-84.90) | - | 87.00 (86.02-87.98) | - |
| Hi-Seg (senior radiologist) | 86.32 (85.63-87.01) | 0.016 | 87.48 (85.25-89.71) | 0.579 |
| Hi-Seg (junior medical student) | 84.73 (83.96-85.50) | 0.045 | 88.07 (86.38-89.75) | 0.682 |
| Hi-Seg (non-medical annotators)[f] | 83.77 (83.16-84.39) | 0.842 | 86.49 (85.15-87.82) | 0.942 |

[a]Dice scores are reported as mean values with 95% confidence intervals (%).

[b]*P*-values were calculated using Wilcoxon signed-rank tests comparing each model with the reference Hi-Seg (overall, five human annotators).

[c]Deep-learning models include U-Net, U-Net++, Attention U-Net, TransUNet and nnU-Net.

[d]Segment Anything Model (SAM) is a prompt-driven foundation model that generates segmentation masks from user-provided prompts, such as points or bounding boxes. HQ-SAM, MedSAM2 and MedSAM are variants of SAM. Pseudo-human prompts (#1-4) were automatically generated from ground-truth masks. SAM, HQ-SAM and MedSAM2 were evaluated using prompts #1-4; “overall” denotes performance averaged across these prompts. MedSAM (bounding box prompt) indicates prompting with a pseudo-human bounding box.

[e]Human-in-the-loop segmentation (Hi-Seg) refers to the proposed framework that integrates real-time interactive prompts provided by human annotators. Annotator groups included one senior radiologist, one junior medical student and three non-medical annotators. Hi-Seg (overall, five human annotators) represents results aggregated across all five annotators, whereas the annotator specified in parentheses denotes the individual performing the interaction (for example, Hi-Seg (senior radiologist)).

[f]Hi-Seg (non-medical annotators) represents the mean performance of three non-medical annotators under human-in-the-loop conditions (Methods).

SAM, Segment Anything Model; HQ-SAM, High-Quality Segment Anything Model; Hi-Seg, human-in-the-loop segmentation.

Table 2. Ablation study of the effects of human prompting and iterative feedback refinement on segmentation performance.

| Model variant | Human prompt[i] | Iterative | Feedback-based refinement | Mean Dice[a] (%) | *P*-value[b] |
|---|---|---|---|---|---|
| Prompt-free SAM[c] | No | No | No | 45.27 | <0.001 |
| One-shot SAM[d] | Pseudo-human | No | No | 65.41 | <0.001 |
| Pseudo-human-prompted SAM (#1, #2)[e] | Pseudo-human | Yes | No | 70.84 | <0.001 |
| Pseudo-human-prompted SAM (#3, #4)[f] | Pseudo-human | No | No | 75.44 | <0.001 |
| Hi-Seg (one-shot)[g] | Real human | No | No | 65.27 | <0.001 |
| **Hi-Seg (full)[h]** | **Real human** | **Yes** | **Yes** | **84.81** | **-** |

[a]Dice scores are reported as averages across the two external test datasets.

[b]*P*-values were calculated using two-sided Wilcoxon signed-rank tests comparing each ablation variant with the reference Hi-Seg (full).

[c]Prompt-free SAM generates segmentation masks in a single pass without any prompts.

[d]One-shot SAM denotes the average segmentation performance of SAM using each pseudo-human prompt (#1-#4), with only one round of prompting, averaged across all pseudo-human prompts.

[e,f]Pseudo-human-prompted SAM (#1, #2) (e) and pseudo-human-prompted SAM (#3, #4) (f) denote the average performance of SAM using pseudo-human prompts #1 and #2, and pseudo-human prompts #3 and #4, respectively.

[g]Hi-Seg (one-shot) denotes segmentation using a single human prompt in a human-in-the-loop setting; values represent the mean performance across five annotators.

[h]Hi-Seg (full) denotes the proposed human-in-the-loop segmentation approach, which integrates three components: human prompting, iterative interaction, and feedback-based refinement; values represent the mean performance across five annotators.

[i]Pseudo-human prompts are automatically generated from ground-truth masks. Real human prompts are generated by annotators during real-time interactive segmentation using Hi-Seg.

SAM is a prompt-driven foundation model that generates segmentation masks from user-provided prompts, such as points or bounding boxes. SAM, Segment Anything Model.

Table 3. Performance and efficiency of human-in-the-loop segmentation compared with manual annotation across annotator groups.

| **Annotator group**[a] | **Annotation mode**[b] | **Dice**[c] **(%)** | **ΔDice**[d] **(%)** | ***P*-value** | **Annotation time per nodule**[f] **(second)** | **ΔTime**[e] **(%)** | ***P*-value**[g] |
|---|---|---|---|---|---|---|---|
| Senior radiologist | Manual | 85.25 (84.6-85.91) | - | - | 14.53 (13.30-15.76) | - | - |
| Senior radiologist | Hi-Seg | 86.47 (85.81-87.14) | +1.43 | <0.001 | 15.03 (14.48-15.59) | +3.46% | <0.001 |
| Junior medical student | Manual | 84.17 (83.47-84.86) | - | - | 25.28 (23.04-27.52) | - | - |
| Junior medical student | Hi-Seg | 85.17 (84.46-85.87) | +1.19 | <0.001 | 17.53 (16.15-18.91) | -30.66% | <0.001 |
| Non-medical annotators[h] | Manual | 81.05 (80.33-81.78) | - | - | 22.16 (21.00-23.32) | - | - |
| Non-medical annotators[h] | Hi-Seg | 84.13 (83.3-84.96) | +3.80 | <0.001 | 16.37 (15.55-17.20) | -26.11% | <0.001 |

[a]Annotator groups included one senior radiologist, one junior medical student, and three non-medical annotators.

[b]Two annotation modes were evaluated: manual segmentation and human-in-the-loop segmentation (Hi-Seg).

[c]Values are presented as means with 95% confidence intervals.

[d,e]ΔDice and ΔTime denote the percentage improvement in Dice score and the percentage reduction in annotation time, respectively, achieved by Hi-Seg relative to manual segmentation.

[f]Annotation time per nodule (seconds) was defined as the cumulative time required to complete segmentation across the axial, coronal, and sagittal planes for each nodule and was averaged across all nodules within each annotator group.

[g]*P*-values were calculated using Wilcoxon signed-rank tests comparing Hi-Seg with manual annotation within each annotator group.

[h]Values represent the mean performance across the three non-medical annotators (Methods).

# Supplementary information for human and AI collaboration for pulmonary nodule segmentation

Supplementary Table 1: Differences in lung nodule characteristics across the three datasets.

| Characteristic | Training (n=1110) | Internal validation (n=275) | Single-center (n=885) | Multi-center (n=134) | *P*-value[b] |
|---|---|---|---|---|---|
| **Size (mm)** | 9.6 (7.4-14.3)[a] | 9.8 (7.5-13.4) | 20.2 (12.9-25.2) | 21.5 (16.7-26.8) | <0.001 |
| **Solid nodule** | 889 (80.1) | 228 (82.9) | 660 (74.6) | 119 (88.8) | <0.001 |
| **Part-solid nodule** | 140 (12.6) | 27 (9,8) | 173 (19.5) | 14 (10.4) | <0.001 |
| **Ground-glass nodule (GGN)** | 81 (7.3) | 20 (7.3) | 52 (5.9) | 1 (0.8) | 0.024 |
| **Spiculation** | 272 (24.5) | 55 (20.0) | 634 (71.6) | 75 (60.0) | <0.001 |
| **Lobulation** | 384 (34.6) | 88 (32.0) | 619 (69.9) | 68 (50.8) | <0.001 |
| **Calcification** | 122 (11.0) | 34 (12.4) | 74 (8.4) | 17(12.7) | 0.099 |
| **Cavitation** | N/A[c] | N/A | 41 (4.6) | 8 (6.0) | 0.500 |
| **Vacuole sign** | N/A | N/A | 41 (4.6) | 13 (9.7) | 0.015 |
| **Pleural tag** | N/A | N/A | 543 (61.4) | 73 (54.5) | 0.129 |
| **Air bronchogram** | N/A | N/A | 179 (20.2) | 20 (14.9) | 0.149 |
| **Fat** | 0 (0.0) | 0 (0.0) | 13 (1.5) | 1 (0.8) | <0.001 |
| **Pseudo cavity** | N/A | N/A | 44 (5.0) | 8 (6.0) | 0.625 |

[a]Values of nodule size are represented as median, 25th, and 75th percentiles; Values of other characteristics are represented as number and percentage.

[b]Continuous variables were analyzed using the Kruskal-Wallis *H* test, and binary variables were analyzed using the $\chi^2$ test with Bonferroni corrections.

[c]N/A indicates the characteristic is not annotated in the corresponding dataset.

Supplementary Table 2: Comparison of segmentation performance of deep learning models on the internal validation dataset in this study with previously reported results.

| Deep learning models[a] | This study[b] | Ref. 23[d] | Ref. 10[e] | Ref. 10[f] | Ref. 10[g] | Ref. 53[h] | Ref. 54[i] |
|---|---|---|---|---|---|---|---|
| **U-Net** | 0.694 ± 0.213[c] | 0.712 ± 0.194 | 0.842 | 0.784 | 0.752 | 0.784 ± 0.144 | 0.691 |
| **U-Net++** | 0.695 ± 0.205 | — | — | — | — | — | 0.725 |
| **Attention U-Net** | 0.681 ± 0.209 | — | 0.858 | 0.795 | 0.761 | 0.786 ± 0.170 | — |
| **TransU-Net** | 0.652 ± 0.202 | — | 0.859 | 0.797 | 0.762 | — | 0.744 |
| **nnU-Net** | 0.785 ± 0.162 | 0.779 ± 0.138 | 0.865 | 0.803 | 0.769 | — | 0.645 |

[a]Five deep learning models evaluated in this study: U-Net, U-Net++, Attention U-Net, TransU-Net and nnU-Net.

[b]In this study, models were evaluated on 256 × 256 two-dimensional images covering the entire lung region from the LIDC-IDRI dataset, with images extracted from axial, coronal and sagittal views.

[c]Dice scores are reported as mean ± standard deviation.

[d]Validation was performed under the same experimental setting as in this study, namely on 256 × 256 axial images covering the entire lung region from the LIDC-IDRI dataset. [e-g]Validation data consisted of 64 × 64 × 64 three-dimensional nodule-centered images; models were externally evaluated on the LIDC-IDRI (e), NSCLC-Radiomics (f) and MosMedData (g) datasets, respectively.

[h]Validation data consisted of 64 × 64 axial images centered on nodules from the LIDC-IDRI dataset.

[i]Validation data consisted of 96 × 96 axial images centered on nodules; models were trained and evaluated on the LNDb dataset.

Our experimental setting in (b) matches that in (g), enabling direct comparison of segmentation performance. “-” indicates that the corresponding results were not reported in the original publications.

Ref. 53: Ni, Y., Xie, Z., Zheng, D., Yang, Y. & Wang, W. Two-stage multitask U-Net construction for pulmonary nodule segmentation and malignancy risk prediction. *Quant. Imaging Med. Surg.* **12**, 292-309 (2022).

Ref. 54: Hu, T., et al. A lung nodule segmentation model based on the transformer with multiple thresholds and coordinate attention. *Sci. Rep.* **14**, 31743 (2024).

Supplementary Table 3: Detailed comparison of segmentation performance between Hi-Seg with real human prompts, deep learning models, and Segment Anything-based methods using pseudo-human prompts across two external test datasets.

| Model | Single-center (n=885) | | Multi-center (n=134) | |
|---|---|---|---|---|
| | Dice (%)[a] | *P*-value[b] | Dice (%)[a] | *P*-value[b] |
| **Deep learning models** | | | | |
| **U-Net** | 70.77 (68.99-72.56) | < 0.001 | 66.52 (61.02-72.01) | < 0.001 |
| **U-Net++** | 69.84 (68.07-71.61) | < 0.001 | 65.25 (59.88-70.62) | < 0.001 |
| **Attention U-Net** | 67.88 (65.96-69.80) | < 0.001 | 63.25 (57.61-68.88) | < 0.001 |
| **TransU-Net** | 63.51 (61.54-65.48) | < 0.001 | 56.89 (51.08-62.70) | < 0.001 |
| **nnU-Net** | 75.42 (73.90-76.94) | < 0.001 | 69.05 (64.21-73.88) | < 0.001 |
| **Segment anything models / Pseudo-human prompts[c]** | | | | |
| **SAM (prompt #1)** | 74.48 (72.95-76.01) | < 0.001 | 81.02 (77.97-84.08) | < 0.001 |
| **SAM (prompt #2)** | 63.73 (62.14-65.32) | < 0.001 | 71.31 (67.55-75.06) | < 0.001 |
| **SAM (prompt #3)** | 55.06 (53.11-57.00) | < 0.001 | 62.30 (57.58-67.03) | < 0.001 |
| **SAM (prompt #4)** | 77.83 (76.68-78.99) | < 0.001 | 80.75 (78.04-83.46) | < 0.001 |
| **HQ-SAM (prompt #1)** | 74.55 (72.94-76.15) | < 0.001 | 82.40 (79.53-85.27) | 0.109 |
| **HQ-SAM (prompt #2)** | 63.44 (61.83-65.05) | < 0.001 | 68.15 (64.26-72.04) | < 0.001 |
| **HQ-SAM (prompt #3)** | 77.04 (75.57-78.51) | < 0.001 | 80.54 (77.25-83.83) | < 0.001 |
| **HQ-SAM (prompt #4)** | 78.95 (77.77-80.12) | < 0.001 | 81.37 (78.72-84.02) | < 0.001 |
| **MedSAM2 (prompt #1)** | 78.26 (76.98-79.54) | < 0.001 | 83.93 (81.99-85.86) | < 0.001 |
| **MedSAM2 (prompt #2)** | 65.19 (63.70-66.69) | < 0.001 | 73.90 (70.52-77.27) | < 0.001 |
| **MedSAM2 (prompt #3)** | 77.96 (76.82-79.10) | < 0.001 | 81.80 (79.74-83.87) | < 0.001 |
| **MedSAM2 (prompt #4)** | 83.08 (82.38-83.78) | < 0.001 | 83.99 (82.48-85.50) | < 0.001 |
| **MedSAM (bounding-box prompt)** | 79.24 (78.76-79.72) | < 0.001 | 81.50 (80.60-82.40) | < 0.001 |
| **Human-in-the-loop segmentation (Hi-Seg)[d]** | | | | |
| **Hi-Seg (senior radiologist)** | 86.32 (85.63-87.01) | <0.001 | 87.48 (85.25-89.71) | <0.001 |
| **Hi-Seg (junior medical student)** | 84.73 (83.96-85.50) | <0.001 | 88.07 (86.38-89.75) | 0.066 |
| **Hi-Seg (non-medical annotators #1)** | 82.78 (81.53-84.02) | 0.525 | 87.46 (85.32-89.60) | 0.014 |
| **Hi-Seg (non-medical annotators #2)** | 81.77 (80.56-82.99) | <0.001 | 84.85 (82.10-87.59) | 0.024 |
| **Hi-Seg (non-medical annotators #3)** | 86.77 (86.21-87.33) | <0.001 | 87.15 (85.11-89.20) | 0.474 |
| **Hi-Seg (overall, five human annotators)[e]** | 84.47 (84.05-84.90) | - | 87.00 (86.02-87.98) | - |

[a]Dice scores quantify the overlap between predicted mask and ground-truth annotation and are reported per case as mean values with 95% confidence intervals.

[b]*P*-values were calculated using Wilcoxon signed-rank tests comparing each method with Hi-Seg (overall, five human annotators).

[c]Segment Anything models include SAM, HQ-SAM, MedSAM2, and MedSAM. Four pseudo-human prompts were automatically generated from ground-truth masks. "SAM (prompt #1)" indicates SAM guided by pseudo-human prompt #1. "MedSAM (bounding box prompt)" indicates MedSAM guided by a pseudo-human bounding box derived from the ground-truth mask (Methods).

[d]Human-in-the-loop segmentation (Hi-Seg) refers to the proposed framework with real-time interactive prompts from human annotators, including one senior radiologist, one junior medical student, and three non-medical annotators. The annotator specified in parentheses indicates who provided the prompt (e.g., Hi-Seg (senior radiologist)).

[e]Hi-Seg (overall, five human annotators) represents aggregated results across all five annotators. SAM, Segment Anything Model; HQ-SAM, High Quality SAM; Hi-Seg, human-in-the-loop segmentation.

Supplementary Table 4a: Segmentation accuracy across nodule characteristic subgroups for human-in-the-loop segmentation (Hi-Seg), manual delineation, and automated methods.

| Nodule characteristic[a] | Hi-Seg Dice[b] (%)[c] | Manual Dice[b] (%)[d] | Pseudo-human-prompted SAMs Dice[b] (%)[e] | Deep learning models Dice[b] (%)[f] | Prompt-free SAM Dice[b] (%)[g] |
|---|---|---|---|---|---|
| **Size** | | | | | |
| **<15 mm (n=275)** | 79.56 (78.67-80.45) | 77.35 (76.59-78.10) | 68.00 (66.69-69.31) | 62.13 (61.27-62.99) | 43.19 (39.19-47.18) |
| **≥15 mm (n=744)** | 86.74 (86.34-87.15) | 84.43 (84.05-84.80) | 69.08 (68.13-70.02) | 77.89 (77.52-78.25) | 46.04 (43.60-48.48) |
| **Density** | | | | | |
| **Solid (*n*=779)** | 86.46 (86.07-86.86) | 83.81 (83.44-84.18) | 69.27 (68.37-70.17) | 76.15 (75.76-76.55) | 50.61 (48.25-52.97) |
| **Part-solid (*n*=187)** | 82.46 (81.65-83.27) | 80.92 (80.23-81.61) | 71.12 (69.60-72.65) | 70.82 (69.98-71.67) | 31.47 (27.16-35.78) |
| **Ground glass nodule (*n*=53)** | 68.72 (65.64-71.79) | 69.08 (66.40-71.77) | 53.43 (49.75-57.12) | 46.56 (44.45-48.66) | 15.49 (9.56-21.42) |
| **Spiculation** | | | | | |
| **w/ (*n*=709)** | 86.23 (85.84-86.63) | 84.01 (83.66-84.36) | 71.35 (70.47-72.24) | 76.12 (75.72-76.52) | 44.49 (42.05-46.93) |
| **w/o (*n*=310)** | 81.54 (80.65-82.43) | 79.10 (78.28-79.91) | 62.92 (61.41-64.43) | 67.95 (67.15-68.75) | 47.05 (43.08-51.02) |
| **Lobulation** | | | | | |
| **w/ (*n*=687)** | 86.18 (85.78-86.58) | 83.84 (83.48-84.20) | 70.96 (70.05-71.87) | 76.08 (75.67-76.48) | 45.00 (42.49-47.51) |
| **w/o (*n*=332)** | 81.96 (81.11-82.81) | 79.77 (79.00-80.54) | 64.30 (62.86-65.73) | 68.58 (67.82-69.34) | 45.84 (42.09-49.58) |
| **Calcification** | | | | | |
| **w/ (*n*=91)** | 84.22 (82.66-85.78) | 82.26 (80.94-83.59) | 63.48 (60.52-66.44) | 73.42 (72.15-74.69) | 41.61 (34.68-48.53) |
| **w/o (*n*=928)** | 84.86 (84.46-85.26) | 82.54 (82.17-82.91) | 69.31 (68.51-70.11) | 73.65 (73.26-74.05) | 45.63 (43.44-47.81) |
| **Cavitation** | | | | | |
| **w/ (*n*=49)** | 88.71 (87.73-89.69) | 86.68 (85.85-87.52) | 65.57 (61.57-69.58) | 78.85 (77.57-80.13) | 47.30 (38.95-55.66) |
| **w/o (*n*=970)** | 84.61 (84.20-85.01) | 82.31 (81.94-82.67) | 68.95 (68.16-69.74) | 73.37 (72.98-73.76) | 45.17 (43.02-47.32) |
| **Vacuole sign** | | | | | |
| **w/ (*n*=54)** | 88.27 (87.32-89.22) | 85.57 (84.67-86.47) | 75.73 (72.99-78.47) | 77.56 (76.19-78.93) | 51.76 (42.78-60.73) |
| **w/o (*n*=965)** | 84.61 (84.20-85.02) | 82.34 (81.98-82.71) | 68.40 (67.60-69.20) | 73.41 (73.03-73.80) | 44.91 (42.76-47.05) |
| **Pleural tag** | | | | | |
| **w/ (*n*=616)** | 86.76 (86.36-87.16) | 84.58 (84.25-84.92) | 72.72 (71.82-73.62) | 76.27 (75.84-76.70) | 44.88 (42.26-47.51) |
| **w/o (*n*=403)** | 81.82 (81.07-82.57) | 79.35 (78.65-80.06) | 62.77 (61.42-64.13) | 69.61 (68.94-70.28) | 45.86 (42.44-49.28) |
| **Air bronchogram** | | | | | |
| **w/ (*n*=199)** | 84.73 (83.98-85.49) | 83.10 (82.46-83.75) | 70.11 (68.49-71.73) | 75.09 (74.36-75.82) | 35.22 (30.82-39.62) |
| **w/o (*n*=820)** | 84.82 (84.38-85.27) | 82.37 (81.96-82.78) | 68.47 (67.59-69.34) | 73.28 (72.85-73.71) | 47.71 (45.38-50.04) |
| **Fat** | | | | | |
| **w/ (*n*=14)** | 85.10 (82.04-88.15) | 82.93 (80.02-85.84) | 66.39 (59.05-73.74) | 73.61 (70.53-76.68) | 68.12 (52.40-83.85) |
| **w/o (*n*=1,005)** | 84.80 (84.41-85.20) | 82.51 (82.15-82.87) | 68.82 (68.04-69.60) | 73.63 (73.26-74.01) | 44.95 (42.86-47.05) |
| **Pseudo cavity** | | | | | |
| **w/ (*n*=52)** | 80.81 (78.89-82.73) | 78.87 (77.07-80.68) | 59.46 (55.86-63.07) | 72.02 (70.48-73.55) | 34.65 (26.72-42.58) |
| **w/o (*n*=967)** | 85.02 (84.62-85.42) | 82.71 (82.35-83.07) | 69.29 (68.50-70.08) | 73.72 (73.34-74.11) | 45.84 (43.69-47.99) |

Supplementary Table 4b: Multiple-comparison analysis of segmentation performance across methods within each nodule subgroup.

| Nodule characteristic[a] | Global test P-value[h] | Hi-Seg[c] | Manual[d] | Pseudo-human-prompted SAMs[e] | Deep learning models[f] | Prompt-free SAM[g] |
|---|---|---|---|---|---|---|
| **Size** | | | | | | |
| **<15 mm (n=275)** | < 0.001 | A[i] | B | C | D | D |
| **≥15 mm (n=744)** | < 0.001 | A | B | C | D | E |
| **Density** | | | | | | |
| **Solid (*n*=779)** | < 0.001 | A | B | C | D | D |
| **Part-solid (*n*=187)** | < 0.001 | A | B | C | D | E |
| **Ground glass nodule (*n*=53)** | < 0.001 | A | A | B | B | C |
| **Spiculation** | | | | | | |
| **w/ (*n*=709)** | < 0.001 | A | B | C | D | E |
| **w/o (*n*=310)** | < 0.001 | A | B | C | C | C |
| **Lobulation** | | | | | | |
| **w/ (*n*=687)** | < 0.001 | A | B | C | D | E |
| **w/o (*n*=332)** | < 0.001 | A | B | C | CD | D |
| **Calcification** | | | | | | |
| **w/ (*n*=91)** | < 0.001 | A | B | C | C | C |
| **w/o (*n*=928)** | < 0.001 | A | B | C | D | E |
| **Cavitation** | | | | | | |
| **w/ (*n*=49)** | < 0.001 | A | B | C | C | C |
| **w/o (*n*=970)** | < 0.001 | A | B | C | D | E |
| **Vacuole sign** | | | | | | |
| **w/ (*n*=54)** | < 0.001 | A | B | BC | C | C |
| **w/o (*n*=965)** | < 0.001 | A | B | C | D | E |
| **Pleural tag** | | | | | | |
| **w/ (*n*=616)** | < 0.001 | A | B | C | D | E |
| **w/o (*n*=403)** | < 0.001 | A | B | C | CD | D |
| **Air bronchogram** | | | | | | |
| **w/ (*n*=199)** | < 0.001 | A | B | C | D | E |
| **w/o (*n*=820)** | < 0.001 | A | B | C | D | D |
| **Fat** | | | | | | |
| **w/ (*n*=14)** | < 0.001 | A | AB | B | B | AB |
| **w/o (*n*=1,005)** | < 0.001 | A | B | C | D | E |
| **Pseudo cavity** | | | | | | |
| **w/ (*n*=52)** | < 0.001 | A | B | C | C | D |
| **w/o (*n*=967)** | < 0.001 | A | B | C | D | E |

[a]Nodule subgroups were stratified according to radiological characteristics. “w/” and “w/o” indicate the presence and absence of a given characteristic, respectively.

[b]Dice scores are reported as the mean (95% confidence interval) across nodules within each subgroup and across the two external datasets (%).

[c]Hi-Seg denotes the proposed human-in-the-loop segmentation approach; values represent the mean performance of five annotators.

[d]Manual denotes human delineation; values represent the mean performance of five annotators.

[e]Pseudo-human-prompted SAMs denote the mean performance of 13 models, including SAM, HQ-SAM, and MedSAM2 with pseudo-human prompts #1-#4, as well as MedSAM prompted by bounding boxes. SAM is a prompt-driven foundation model that generates segmentation masks from user-provided prompts, such as points or bounding boxes. HQ-SAM, MedSAM2 and MedSAM are variants of SAM.

[f]Deep-learning models denote the mean performance of U-Net, U-Net++, Attention U-Net, TransU-Net, and nnU-Net.

[g]Prompt-free SAM generates masks in a single pass without any prompts.

[h]Global text *P* values were derived from the Friedman test.

[i]Letters A-E indicate ranking by descending mean Dice score. Columns sharing the same letter are not significantly different at a significance level of 0.05. P-values of pairwise comparisons were

adjusted using Dunn & Šidák’s post hoc test (Methods). SAM, Segment Anything Model.

Supplementary Table 5: Segmentation efficiency of human annotators using Hi-Seg or manual annotation across pulmonary nodule characteristic subgroups.

| Nodule characteristic[a] | Hi-Seg[b] (senior radiologist) | Manual[c] (senior radiologist) | Hi-Seg[d] (junior medical student) | Manual[e] (junior medical student) | Hi-Seg[f] (non-medical annotators) | Manual[g] (non-medical annotators) |
|---|---|---|---|---|---|---|
| **Size** | | | | | | |
| **<15 mm (n=275)** | 14.12(13.18-15.06) | 10.06(7.98-12.14) | 19.38(16.67-22.08) | 14.47(11.88-17.06) | 14.39(13.65-15.13) | 16.91(16.07-17.74) |
| **≥15 mm (n=744)** | 15.38(14.70-16.05) | 16.19(14.69-17.68) | 16.89(15.31-18.47) | 29.23(26.38-32.09) | 16.53(15.90-17.15) | 21.59(20.87-22.30) |
| **Density** | | | | | | |
| **Solid (*n*=779)** | 14.20(13.60-14.81) | 14.34(12.93-15.75) | 16.60(15.09-18.10) | 25.21(22.72-27.69) | 14.96(14.43-15.49) | 20.70(20.03-21.36) |
| **Part-solid (*n*=187)** | 17.04(15.70-18.38) | 14.29(12.61-15.98) | 20.52(16.56-24.48) | 28.06(21.87-34.24) | 19.05(17.58-20.52) | 19.56(18.21-20.91) |
| **Ground glass nodule (*n*=53)** | 20.21(17.36-23.06) | 18.16(8.06-28.25) | 21.22(18.76-23.68) | 15.17(10.41-19.94) | 19.59(17.79-21.39) | 17.56(15.64-19.48) |
| **Spiculation** | | | | | | |
| **w/ (*n*=709)** | 15.14(14.49-15.80) | 15.21(13.70-16.72) | 16.85(15.18-18.51) | 26.82(23.99-29.65) | 16.23(15.62-16.84) | 21.17(20.42-21.92) |
| **w/o (*n*=310)** | 14.79(13.76-15.82) | 12.99(10.85-15.12) | 19.19(16.82-21.56) | 21.60(18.14-25.05) | 15.30(14.43-16.17) | 18.39(17.63-19.15) |
| **Lobulation** | | | | | | |
| **w/ (*n*=687)** | 15.24(14.56-15.92) | 15.11(13.70-16.53) | 16.60(14.92-18.29) | 26.95(24.02-29.88) | 16.15(15.53-16.76) | 20.81(20.11-21.51) |
| **w/o (*n*=332)** | 14.62(13.66-15.58) | 13.32(10.92-15.73) | 19.54(17.22-21.85) | 21.65(18.49-24.81) | 15.54(14.67-16.40) | 19.32(18.31-20.32) |
| **Calcification** | | | | | | |
| **w/ (*n*=91)** | 17.19(14.54-19.85) | 16.55(12.88-20.21) | 23.09(15.28-30.89) | 31.52(23.35-39.70) | 15.90(13.84-17.97) | 20.26(18.63-21.90) |
| **w/o (*n*=928)** | 14.83(14.28-15.38) | 14.33(13.02-15.64) | 17.01(15.72-18.30) | 24.60(22.29-26.92) | 15.95(15.44-16.46) | 20.33(19.72-20.94) |
| **Cavitation** | | | | | | |
| **w/ (*n*=49)** | 17.82(15.20-20.44) | 21.18(14.71-27.65) | 15.81(12.84-18.79) | 33.97(23.48-44.46) | 18.29(15.62-20.96) | 23.64(19.55-27.72) |
| **w/o (*n*=970)** | 14.90(14.33-15.46) | 14.19(12.94-15.45) | 17.65(16.22-19.07) | 24.79(22.51-27.07) | 15.83(15.32-16.34) | 20.16(19.59-20.72) |
| **Vacuole sign** | | | | | | |
| **w/ (*n*=54)** | 13.83(11.77-15.89) | 11.83(8.64-15.02) | 13.97(12.25-15.69) | 32.08(19.57-44.60) | 14.11(12.74-15.47) | 23.01(19.36-26.67) |
| **w/o (*n*=965)** | 15.10(14.53-15.68) | 14.68(13.39-15.97) | 17.75(16.32-19.19) | 24.85(22.60-27.11) | 16.05(15.53-16.57) | 20.17(19.60-20.74) |
| **Pleural tag** | | | | | | |
| **w/ (*n*=616)** | 15.17(14.46-15.88) | 14.91(13.33-16.48) | 17.03(15.06-19.01) | 26.11(23.10-29.12) | 15.96(15.32-16.60) | 21.25(20.45-22.06) |
| **w/o (*n*=403)** | 14.84(13.95-15.73) | 13.95(11.95-15.95) | 18.36(16.68-20.04) | 23.87(20.60-27.14) | 15.94(15.14-16.74) | 18.91(18.15-19.66) |
| **Air bronchogram** | | | | | | |
| **w/ (*n*=199)** | 17.65(16.31-18.98) | 17.83(15.35-20.30) | 18.36(15.29-21.43) | 32.87(26.53-39.22) | 18.99(17.76-20.21) | 20.84(19.60-22.08) |
| **w/o (*n*=820)** | 14.40(13.80-15.00) | 13.73(12.33-15.14) | 17.36(15.84-18.89) | 23.37(21.08-25.67) | 15.22(14.67-15.76) | 20.20(19.55-20.85) |
| **Fat** | | | | | | |
| **w/ (*n*=14)** | 16.39(5.20-27.59) | 19.09(4.46-33.71) | 19.40(5.50-33.30) | 23.37(9.91-36.84) | 12.96(9.54-16.39) | 18.80(13.08-24.52) |
| **w/o (*n*=1,005)** | 15.02(14.47-15.56) | 14.47(13.23-15.71) | 17.53(16.16-18.90) | 25.25(22.99-27.50) | 15.99(15.49-16.50) | 20.34(19.77-20.92) |
| **Pseudo cavity** | | | | | | |
| **w/ (*n*=52)** | 21.07(17.68-24.46) | 21.26(17.15-25.37) | 17.60(15.08-20.12) | 37.64(27.24-48.05) | 22.15(19.08-25.23) | 22.46(18.99-25.93) |
| **w/o (*n*=967)** | 14.71(14.16-15.26) | 14.17(12.89-15.45) | 17.55(16.12-18.99) | 24.57(22.29-26.85) | 15.62(15.12-16.12) | 20.21(19.63-20.78) |
| **Single center dataset (n=885)** | 14.77(14.18-15.36) | 14.67(13.28-16.07) | 16.96(15.42-18.49) | 23.16(20.81-25.51) | 15.78(15.29-16.26) | 20.10(19.50-20.69) |
| **Multi center dataset (n=134)** | 16.80(15.21-18.39) | 13.58(11.70-15.46) | 21.51(19.29-23.73) | 38.89(32.50-45.29) | 17.08(15.05-19.11) | 21.85(19.99-23.70) |
| **Overall (n=1,019)** | 15.04(14.48-15.59) | 14.53(13.30-15.76) | 17.56(16.19-18.92) | 25.22(22.99-27.45) | 15.95(15.45-16.45) | 20.32(19.75-20.90) |

Segmentation efficiency (seconds per nodule) was defined as the cumulative segmentation time across axial, coronal and sagittal planes and is reported as mean values with 95% confidence intervals.

[a]Nodule subgroups were stratified by radiological characteristics; "w/" and "w/o" denote presence or absence.

[b-e]Hi-Seg denotes the human-in-the-loop segmentation framework; Hi-Seg and Manual (senior radiologist/junior medical student) indicate segmentation by the corresponding annotator using Hi-Seg or manual delineation, respectively.

[f,g]Hi-Seg and Manual (non-medical annotators) denote mean segmentation efficiency across three non-medical annotators using Hi-Seg or manual delineation, respectively. Hi-Seg, human-in-the-loop segmentation.

Supplementary Table 6: Failure rates of different methods for each nodule subgroup.

| Nodule characteristic[a] | Hi-Seg (senior radiologist)[b] | Manual (senior radiologist)[c] | Hi-Seg (junior medical student)[d] | Manual (junior medical student)[e] | Hi-Seg (non-medical annotators)[f] | Manual (non-medical annotators)[g] | Deep learning models[h] | Prompt-free SAM[i] |
|---|---|---|---|---|---|---|---|---|
| **Overall (n=1,019)** | 1.28 | 1.18 | 1.86 | 1.96 | 3.7 | 3.47 | 21.41 | 54.47 |
| **Size** | | | | | | | | |
| **<15 mm (n=275)** | 2.55 | 1.82 | 3.64 | 3.27 | 6.42 | 6.18 | 19.64 | 53.82 |
| **≥15 mm (n=744)** | 0.81 | 0.94 | 1.21 | 1.48 | 2.69 | 2.46 | 22.07 | 54.7 |
| **Density** | | | | | | | | |
| **Solid (*n*=779)** | 1.41 | 1.41 | 1.41 | 1.41 | 2.35 | 2.4 | 20.69 | 47.37 |
| **Part-solid (*n*=187)** | 0 | 0 | 1.07 | 1.6 | 4.1 | 3.21 | 18.61 | 74.87 |
| **Ground glass nodule (*n*=53)** | 3.77 | 1.89 | 11.32 | 11.32 | 22.01 | 20.13 | 41.89 | 86.79 |
| **Spiculation** | | | | | | | | |
| **w/ (*n*=709)** | 0.71 | 0.56 | 1.13 | 0.99 | 2.44 | 2.07 | 18.67 | 56.56 |
| **w/o (*n*=310)** | 2.58 | 2.58 | 3.55 | 4.19 | 6.56 | 6.67 | 27.68 | 49.68 |
| **Lobulation** | | | | | | | | |
| **w/ (*n*=687)** | 0.87 | 0.44 | 1.31 | 1.46 | 2.18 | 1.99 | 18.86 | 55.46 |
| **w/o (*n*=332)** | 2.11 | 2.71 | 3.01 | 3.01 | 6.83 | 6.53 | 26.69 | 52.41 |
| **Calcification** | | | | | | | | |
| **w/ (*n*=91)** | 1.1 | 1.1 | 3.3 | 3.3 | 4.76 | 3.66 | 27.69 | 59.34 |
| **w/o (*n*=928)** | 1.29 | 1.19 | 1.72 | 1.83 | 3.59 | 3.45 | 20.8 | 53.99 |
| **Cavitation** | | | | | | | | |
| **w/ (*n*=49)** | 0 | 0 | 0 | 0 | 0.68 | 0 | 26.94 | 55.1 |
| **w/o (*n*=970)** | 1.34 | 1.24 | 1.96 | 2.06 | 3.85 | 3.64 | 21.13 | 54.43 |
| **Vacuole sign** | | | | | | | | |
| **w/ (*n*=54)** | 0 | 0 | 0 | 0 | 0.62 | 0.62 | 15.56 | 50 |
| **w/o (*n*=965)** | 1.35 | 1.24 | 1.97 | 2.07 | 3.87 | 3.63 | 21.74 | 54.72 |
| **Pleural tag** | | | | | | | | |
| **w/ (*n*=616)** | 0.16 | 0.65 | 0.81 | 1.14 | 2.11 | 1.57 | 17.08 | 56.98 |
| **w/o (*n*=403)** | 2.98 | 1.99 | 3.47 | 3.23 | 6.12 | 6.37 | 28.04 | 50.62 |
| **Air bronchogram** | | | | | | | | |
| **w/ (*n*=199)** | 0 | 0 | 1.01 | 1.01 | 2.68 | 2.01 | 20 | 69.35 |
| **w/o (*n*=820)** | 1.59 | 1.46 | 2.07 | 2.2 | 3.94 | 3.82 | 21.76 | 50.85 |
| **Fat** | | | | | | | | |
| **w/ (*n*=14)** | 0 | 0 | 0 | 0 | 2.38 | 2.38 | 22.86 | 21.43 |
| **w/o (*n*=1,005)** | 1.29 | 1.19 | 1.89 | 1.99 | 3.71 | 3.48 | 21.39 | 54.93 |
| **Pseudo cavity** | | | | | | | | |
| **w/ (*n*=52)** | 0 | 0 | 5.77 | 7.69 | 7.05 | 7.05 | 35 | 71.15 |
| **w/o (*n*=967)** | 1.34 | 1.24 | 1.65 | 1.65 | 3.52 | 3.27 | 20.68 | 53.57 |

Note: Failure rate is defined as the percentage of failure cases relative to the total number of nodules within each subgroup (%). A failure case is defined as an image with a model Dice score ≤50%[52].

[a]Nodule subgroups were stratified according to radiological characteristics. "w/" and "w/o" indicate the presence and absence of a given characteristic, respectively.

[b-e]Hi-Seg denotes the proposed human-in-the-loop segmentation framework; Hi-Seg (senior radiologist/junior medical student) and Manual (senior radiologist/junior medical student) denote segmentation performed by the corresponding annotator using Hi-Seg or manual delineation, respectively.

[f,g]Hi-Seg (non-medical annotators) and Manual (non-medical annotators) denote the mean segmentation performance of three non-medical annotators using Hi-Seg or manual delineation, respectively.

[h]Deep learning models represent the mean performance across five models: U-Net, U-Net++, Attention U-Net, TransUNet, and nnU-Net.

[i]Prompt-free SAM generates segmentation masks in a single pass without any prompts.

Hi-Seg, human-in-the-loop segmentation; SAM, Segment Anything Model.

## Algorithm for 2D image and mask extraction from a 3D CT image and its corresponding mask

The algorithm is designed to extract 2D slices in the axial, coronal, and sagittal planes, centered around a given nodule, with appropriate cropping and padding.

---

**Algorithm 1.**

---

**Input**: resampled 3D CT image $\boldsymbol{I}_{\mathrm{3D}}$, resampled 3D mask $\boldsymbol{M}_{\mathrm{3D}}$, scaled nodule center $(x_0, y_0, z_0)$

**Output**: cropped 2D images $\boldsymbol{I}_{\mathrm{axial}}$, $\boldsymbol{I}_{\mathrm{coronal}}$, $\boldsymbol{I}_{\mathrm{sagittal}}$ and corresponding ground-truth masks $\boldsymbol{M}_{\mathrm{axial}}$, $\boldsymbol{M}_{\mathrm{coronal}}$, $\boldsymbol{M}_{\mathrm{sagittal}}$

1: **for** $v$ in {axial, coronal, sagittal} **do**
    // Extract 2D slices along each plane. Initialize the cropping center.
2:   **if** $v$ = axial **then**
3:     $\boldsymbol{I} \leftarrow \boldsymbol{I}_{\mathrm{3D}}[:,:,z_0]$;
4:     $\boldsymbol{M} \leftarrow \boldsymbol{M}_{\mathrm{3D}}[:,:,z_0]$;
5:     $(a, b) \leftarrow (256, 256)$; \\ coordinates of the initial cropping center
6:     $(a_0, b_0) \leftarrow (x_0, y_0)$; \\ coordinates of the nodule center on the 2D plane
7:   **else if** $v$ = coronal **then**
8:     $\boldsymbol{I} \leftarrow \boldsymbol{I}_{\mathrm{3D}}[:, y_0, :]$;
9:     $\boldsymbol{M} \leftarrow \boldsymbol{M}_{\mathrm{3D}}[:, y_0, :]$;
10:     $(a, b) \leftarrow (256, z_0 + \delta)$; \\ $\delta \sim U[-20, 20]$ is a random disturbance
11:     $(a_0, b_0) \leftarrow (x_0, z_0)$;
12:   **else**
13:     $\boldsymbol{I} \leftarrow \boldsymbol{I}_{\mathrm{3D}}[x_0, :, :]$;
14:     $\boldsymbol{M} \leftarrow \boldsymbol{M}_{\mathrm{3D}}[x_0, :, :]$;
15:     $(a, b) \leftarrow (256, z_0 + \delta)$; \\ $\delta \sim U[-20, 20]$ is a random disturbance
16:     $(a_0, b_0) \leftarrow (y_0, z_0)$;
17:   **end if**
    \\ If the nodule cannot be fully included in the cropped patch, move the cropping center toward the nodule center by 25 pixels
18:   **while** $\exists (a', b')\ s.t.\ \boldsymbol{M}(a', b') = 1$ **and** ($a' \notin [a-127, a+128]$ **or** $b' \notin [b-127, b+128]$) **do**
19:     $\boldsymbol{\Delta} \leftarrow (a_0 - a', b_0 - b')$;
20:     $(a, b) \leftarrow (a, b) + 25\boldsymbol{\Delta}/\|\boldsymbol{\Delta}\|_2$;
21:   **end while**
    \\ Pad pixels around the image with zeros (128 pixels at each direction), in case of the cropped range exceeds the original image size.
22:   $\boldsymbol{I}_v \leftarrow \mathbf{zeros}(\mathrm{width}(\boldsymbol{I}) + 256, \mathrm{height}(\boldsymbol{I}) + 256)$; // zero matrix
23:   $\boldsymbol{M}_v \leftarrow \mathbf{zeros}(\mathrm{width}(\boldsymbol{I}) + 256, \mathrm{height}(\boldsymbol{I}) + 256)$
24:   $\boldsymbol{I}_v[129{:}128 + \mathrm{width}(\boldsymbol{I}), 129{:}128 + \mathrm{height}(\boldsymbol{I})] \leftarrow \boldsymbol{I}$;
25:   $\boldsymbol{M}_v[129{:}128 + \mathrm{width}(\boldsymbol{I}), 129{:}128 + \mathrm{height}(\boldsymbol{I})] \leftarrow \boldsymbol{M}$;
    \\Crop the $256 \times 256$ 2D image and ground-truth mask
26:   $\boldsymbol{I}_v \leftarrow \boldsymbol{I}_v[a-127{:}a+128, b-127{:}b+128]$;
27:   $\boldsymbol{M}_v \leftarrow \boldsymbol{M}_v[a-127{:}a+128, b-127{:}b+128]$;
28: **end for**
29: Return $\boldsymbol{I}_{\mathrm{axial}}$, $\boldsymbol{I}_{\mathrm{coronal}}$, $\boldsymbol{I}_{\mathrm{sagittal}}$, $\boldsymbol{M}_{\mathrm{axial}}$, $\boldsymbol{M}_{\mathrm{coronal}}$, $\boldsymbol{M}_{\mathrm{sagittal}}$

---